\documentclass[journal]{IEEEtran}

\usepackage{ifpdf}
\usepackage{cite}
\usepackage{amsmath}
\usepackage[mathcal]{eucal}

\usepackage{gensymb}
\usepackage{epsfig} 
\usepackage{algorithm}   
\usepackage{algpseudocode} 
\usepackage{adjustbox}
\usepackage{array}
\usepackage{mathtools}
\usepackage{colortbl}
\usepackage{commath}
\usepackage{mdwmath}
\usepackage{eqparbox}
\usepackage{booktabs}
\usepackage{url}
\usepackage{multirow,array}
\usepackage{graphicx}
\usepackage{bm}
\usepackage{textcomp} 
\usepackage{amssymb} 
\usepackage{threeparttable}

\RequirePackage{ifpdf}
\ifpdf \PassOptionsToPackage{colorlinks=true}{hyperref} \fi
\RequirePackage{hyperref}
\hypersetup{linkcolor=black,citecolor=green}

\ifCLASSOPTIONcompsoc
\usepackage[caption=false, font=normalsize, labelfont=sf, textfont=sf]{subfig}
\else
\usepackage[caption=false, font=footnotesize]{subfig}
\fi

\graphicspath{{./results/}}
\DeclareGraphicsExtensions{.pdf,.jpeg,.png,.jpg}
   
\IEEEoverridecommandlockouts                             
\usepackage{graphics} 
\usepackage{epsfig} 
\usepackage{mathptmx} 
\usepackage{times} 
\usepackage{mdwmath}

\DeclareMathOperator*{\argmin}{arg\,min}

\begin{document}
\renewcommand{\arraystretch}{1.3}
\title{RoSLAC: Robust Simultaneous Localization and Calibration of Multiple Magnetometers}

\author{

Qiyang Lyu$^{*}$, Zhenyu Wu$^{*}$, Wei Wang, Hongming Shen, and Danwei Wang,~\IEEEmembership{Life Fellow,~IEEE}

\vspace{-1em}

\thanks{ 

This research is supported by the National Research Foundation, Singapore under its Medium Sized Centre for Advanced Robotics Technology Innovation (CARTIN). (\textit{Corresponding author: Danwei Wang})

The authors are with the School of Electrical and Electronic Engineering, Nanyang Technological University, Singapore 639798 (e-mail: \{qiyang001, zhenyu002, wei013\}@e.ntu.edu.sg; \{hongming.shen, edwwang\}@ntu.edu.sg).

$^{*}$Co-first authorship
}}


\maketitle

\begin{abstract}
Localization of autonomous mobile robots (AMRs) in enclosed or semi-enclosed environments such as offices, hotels, hospitals, indoor parking facilities, and underground spaces where GPS signals are weak or unavailable remains a major obstacle to the deployment of fully autonomous systems. Infrastructure-based localization approaches, such as QR codes and RFID, are constrained by high installation and maintenance costs as well as limited flexibility, while onboard sensor-based methods, including LiDAR- and vision-based solutions, are affected by ambiguous geometric features and frequent occlusions caused by dynamic obstacles such as pedestrians. Ambient magnetic field (AMF)-based localization has therefore attracted growing interest in recent years because it does not rely on external infrastructure or geometric features, making it well suited for AMR applications such as service robots and security robots. However, magnetometer measurements are often corrupted by distortions caused by ferromagnetic materials present on the sensor platform, which bias the AMF and degrade localization reliability. As a result, accurate magnetometer calibration to estimate distortion parameters becomes essential. Conventional calibration methods that rely on rotating the magnetometer are impractical for large and heavy platforms. To address this limitation, this paper proposes a robust simultaneous localization and calibration (\textit{RoSLAC}) approach based on alternating optimization, which iteratively and efficiently estimates both the platform pose and magnetometer calibration parameters. Extensive evaluations conducted in high-fidelity simulation and real-world environments demonstrate that the proposed \textit{RoSLAC} method achieves high localization accuracy while maintaining low computational cost compared with state-of-the-art magnetometer calibration techniques.
\end{abstract}

\section{Introduction}
Accurate localization is a fundamental capability for autonomous mobile robots (AMRs)
~\cite{Wu2024MGLT, its_fusion_review}. 
Acting as the perceptual foundation, localization enables AMRs to transform sensory observations into positions and orientations within a predefined coordinate frame, allowing them to understand the environment. Typical tasks following localization include pose tracking\cite{vins, fastlio2}, loop closure\cite{hess2016real}, and navigation\cite{fuse_sensor_navigation}. To achieve reliable localization, various sensing modalities have been explored. Geometric approaches often rely on extracting and matching features such as corners, edges, and planes captured by cameras or LiDARs. Other techniques utilize signal intensity or phase differences from active beacons\cite{Poulose2020UWB} or GNSS\cite{Cho2024Outdoor}, or read encoded information from pre-deployed infrastructures such as QR codes\cite{nazemzadeh2017indoor}, magnetic tapes\cite{Jiang2021novel}, or reflectors\cite{roos2021mobile}. However, these existing methods can fail under challenging conditions. On one hand, feature-based geometric approaches are highly sensitive to feature quality and can degrade in environments with repetitive or ambiguous structures including long corridors or large industrial warehouses. GNSS will also drift significantly in indoor environments. On the other hand, beacon-based systems are time-consuming, labor-intensive, and deployment and maintenance-costly\cite{wu2020infrastructure}. Thus, there remains an urgent need for an alternative source of information that provides stable, unique, and infrastructure-free localization cues.

Inspired by the ability of birds to navigate using the Earth’s magnetic field\cite{mouritsen2018long}, recent studies have explored ambient magnetic fields (AMF) as a novel localization modality\cite{kok2018scalable, viset2022ekf, Wu2024MGLT, shenIDFMFLInfrastructurefreeDriftfree2024a}. Although the Earth’s magnetic field is locally uniform and only supports kilometer-level localization, it becomes spatially distinctive in built environments due to distortions caused by ferromagnetic structures such as reinforcing steel bars, shelves, and furniture. These distortions create local magnetic anomalies that are both temporally stable and spatially unique\cite{li2012feasible}. As a result, AMF can serve as a pervasive and re-identifiable signature for localization through magnetic pattern matching, with accuracy reaching the centimeter level. Moreover, unlike visual or LiDAR features, magnetic features can capture structures hidden behind walls or floors and require no additional infrastructure.

A major practical challenge of magnetic localization on AMRs, however, is the interference caused by ferromagnetic components within the robot itself\cite{sieblerMagneticFieldbasedIndoor2023}. Steel chassis and onboard actuators can dynamically distort the measured magnetic field, making raw magnetometer readings unreliable and platform-dependent\cite{kanamori1960crystal}. Some magnetic localization methods\cite{robertson2013simultaneous, kok2018scalable, shi2022PDR} assume ideal sensors and non-magnetic carriers, which is reasonable for body-mounted human sensing but rarely valid for AMRs. To alleviate this issue, pre-calibration is typically performed before localization. In such procedures, the entire robot-sensor system must be manually rotated in all directions to collect sufficiently informative data for observable calibration\cite{vasconcelosGeometricApproachStrapdown2011}. However, this requirement is practical only for small and lightweight devices, and becomes infeasible for large or heavy AMRs.

To address this limitation, our previous work\cite{lyu2025l2mcalibonekeycalibrationmethod} proposed a one-key calibration method based on a pre-built magnetic anomaly map. Instead of rotating the AMR, the method estimates calibration parameters in a single run by moving the system through a known spatially varying AMF with auxiliary odometry. However, it still operates offline and requires post-processing. In this paper, we go one step further and propose an online magnetometer calibration framework that simultaneously compensates carrier-induced magnetic distortions and localizes the system in real time within the magnetic map. The main contributions are summarized as follows:
\begin{itemize}
	\item A simultaneous localization and calibration framework for magnetometers is proposed, which is feasible for online operation and robust to poor parameter initialization.
	\item A novel alternating optimization approach is proposed to ensure both accuracy and computational efficiency in localization and calibration.
	\item Extensive simulations and real-world experiments validate the effectiveness of the proposed method in accurately estimating calibration parameters and enhancing localization performance.
\end{itemize}

The remainder of this paper is organized as follows. Section \ref{sec:lr} reviews related work. Section \ref{sec:model} describes the system model. Section \ref{sec:method} presents the proposed method. Section \ref{sec:exp} evaluates the approach through simulation and real-world experiments. Section~\ref{sec:ablation} explains the necessity of each module. Finally, Section \ref{sec:conclude} concludes the paper.

\section{Literature Review} \label{sec:lr}
The literature review is organized in two aspects: magnetic calibration methods and the magnetic  field-based localization.
\vspace{-2em}
\subsection{Magnetometer Calibration Methods}
Magnetic calibration estimates parameters that model distortions in the measured magnetic field and correct magnetometer readings. These distortions are commonly divided into two categories. \textbf{(a) Interior distortion} arises from sensor imperfections, including axis scaling errors, bias, and axis non-orthogonality~\cite{yangCorrectionMethodThreeAxis2022}. Interior calibration aims to compensate for these intrinsic effects. \textbf{(b) Exterior distortion} is caused by ferromagnetic materials on the carrier platform. It is commonly modeled by the Tolles-Lawson (T-L) model~\cite{tollesCompensationAircraftMagnetic1954}, which represents the disturbance as a \(3 \times 4\) affine transformation with a scaling matrix and a bias vector. Exterior calibration therefore seeks to estimate this transformation to recover the undistorted magnetic field. Existing magnetometer calibration methods can be broadly grouped into two categories.

\paragraph{Magnetometer Offline Calibration} Offline calibration is the most widely studied approach. It is typically performed in a static and homogeneous magnetic field by rotating the magnetometer through a sufficiently diverse set of orientations. Under such conditions, the magnetic field magnitude is direction-independent, so ideal measurements in the magnetometer frame should lie on a sphere centered at the origin. Calibration therefore aims to estimate a transformation that maps distorted measurements onto this sphere. Representative methods include Singular Value Decomposition (SVD)~\cite{vasconcelosGeometricApproachStrapdown2011} and Least Squares (LS) fitting~\cite{chiCalibrationTriaxialMagnetometer2019}. To improve robustness, prior work has incorporated field uncertainty through Total Least Squares (TLS)~\cite{chenMagneticFieldInterference2021} and addressed ill-conditioned measurements using regularized LS~\cite{ningCompensationTechnologyVehicle2024}. More recently, \cite{lyu2025l2mcalibonekeycalibrationmethod} extended offline calibration to non-homogeneous fields by using an auxiliary sensor for localization within a pre-built magnetic map.

\paragraph{Magnetometer Online Calibration} Although offline calibration is generally robust to initialization and can achieve high accuracy, it is often computationally demanding, time-consuming, and dependent on carefully collected data. As a result, increasing attention has been given to online calibration methods, which are typically supported by additional sensing modalities. For instance, Han et al.~\cite{hanExtendedKalmanFilterBased2017} used an Extended Kalman Filter (EKF) to align magnetometer measurements with gyroscope's motion data, thereby improving observability along weakly excited directions. More recently, Siebler et al.~\cite{sieblerMagneticFieldbasedIndoor2023} extended online calibration to unmanned ground vehicles by proposing a Rao-Blackwellized Particle Filter (RBPF) that jointly estimates vehicle pose and calibration parameters through a decomposition of the full state into linear and nonlinear components.
\vspace{-1em}
\subsection{Magnetic Field-based Localization Methods}
Given calibrated magnetometers, magnetic-based localization systems can be constructed. Magnetic information initially serves as an additional sensing modality fused into an existing localization framework. Robertson et al.~\cite{robertson2013simultaneous} proposed MagSLAM, which uses a hierarchical grid map and performs localization by matching local magnetic field distributions to grid cells. Similarly, Kok et al.~\cite{kok2018scalable} introduced hexagonal block tiling combined with reduced-rank Gaussian Process regression and RBPF to enable localization using smartphone magnetometers, assisted by pedestrian dead reckoning (PDR). Shi et al.~\cite{shi2022PDR} further employed particle filtering to fuse PDR and magnetic information for localization on a pre-built map. To improve real-time performance, Viset et al.~\cite{viset2022ekf} replaced RBPF with an Extended Kalman Filter (EKF). In Wang et al.'s work~\cite{wang2017exponentially}, an enhanced exponentially weighted particle filter was proposed to reduce the computational complexity.

While the above methods rely on fusion with wheel odometry~\cite{wang2017exponentially}, PDR~\cite{kok2018scalable,shi2022PDR}, or IMU measurements~\cite{robertson2013simultaneous,viset2022ekf}, other studies focus on purely magnetic-based localization. In~\cite{yeh2020study}, a $k$-Nearest Neighbors (KNN) approach was used to match magnetometer measurements to a magnetic map. Later, Wu et al.~\cite{wu2023global} addressed the limitation of orientation dependency in earlier methods by introducing a maximum a posteriori (MAP) estimator and an HVT rotation-invariant descriptor, enabling orientation-independent localization.
 
However, most of the aforementioned approaches rely on a single magnetometer, which limits the observability of magnetic features. Similar to visual- or LiDAR-based SLAM systems~\cite{vins,fastlio2}, such methods may suffer from degeneracy when the magnetic field is locally flat. To mitigate this issue, Subbu et al.~\cite{Subbu2013LocateMe} proposed magnetic sequence matching using dynamic time warping (DTW) to enhance observability for indoor pedestrian localization. In Wu et al.'s work~\cite{Wu2024MGLT}, a sliding magnetometer mounted on an aluminum bar was used to accumulate magnetic sequences for localization initialization. This idea was further extended in~\cite{shenIDFMFLInfrastructurefreeDriftfree2024a}, where a magnetometer bar was employed and the problem was formulated as a stochastic optimization task, improving purely magnetic-based localization accuracy by addressing heavy-tailed noise effects.

In summary, although both magnetometer calibration and magnetic-based localization have been extensively studied, the literature on jointly performing online calibration and localization remains limited. This increases the operational complexity, which requires calibration to be completed prior to localization. This paper therefore focuses on addressing this limitation by investigating a unified framework for simultaneous localization and calibration.
\vspace{-1em}

\section{System Modeling} \label{sec:model}
\subsection{Magnetic Field-based Robot Localization Modeling}
This paper considers a tri-axial magnetometer array for localization by matching sensor measurements to a pre-built magnetic map $\mathcal{M}(\cdot)$, defined as a dense grid of magnetic values over a predefined area. At time $t_k$, the magnetic measurements are denoted by $\mathcal{B}_{t_k} = \{\mathbf{B}_{t_k}^i\}_{i=1}^N$, where $\mathbf{B}_{t_k}^i = [B_{x,t_k}^i,\; B_{y,t_k}^i,\; B_{z,t_k}^i]^\top$ is the reading of the $i$-th magnetometer, and $N$ is the number of sensors in the array. The robot pose in the map frame at time $t_k$ is defined as $\mathbf{x}_{t_k} = [\mathbf{p}_{t_k}^\top,\; \boldsymbol{\phi}_{t_k}^\top]^\top$, where $\mathbf{p}_{t_k} \in \mathbb{R}^3$ and $\boldsymbol{\phi}_{t_k} \in \mathbb{R}^3$ denote the robot position and orientation, respectively. Given $\mathbf{x}_{t_k}$, the pose of each magnetometer in the array can be expressed in the map frame as:
\begin{equation}
		\small{
		\label{eq:non_seq_loc}
		\begin{aligned}
			f^i(\mathbf{x}_{t_k}) &:= \left(\mathbf{R}_{t_k}^i,\; \mathbf{p}_{t_k}^i\right)
			= \left(
			\exp\!\left(\boldsymbol{\phi}_{t_k}^{\wedge}\right)\, {^b}\mathbf{R}_m^i,\;
			\exp\!\left(\boldsymbol{\phi}_{t_k}^{\wedge}\right)\, {^b}\mathbf{p}_m^i + \mathbf{p}_{t_k}
			\right)
		\end{aligned}
	}
\end{equation}
where ${^b}\mathbf{R}_m^i$ and ${^b}\mathbf{p}_m^i$ denote the extrinsic rotation and translation between the $i$-th magnetometer frame and the robot body frame, respectively. Assuming that the magnetic map is accurate, the magnetic field at a given sensor position $\mathbf{p}$ within the mapped region is defined as
\begin{equation}
			\mathbf{B}_{\text{map}} = \mathcal{M}(\mathbf{p})
\end{equation}
where $\mathcal{P} = \{\mathbf{p} \mid \mathbf{p} \in \mathbb{R}^3 \ \text{and within the mapped region}\}$ and $\mathcal{M}: \mathcal{P} \subset \mathbb{R}^3 \rightarrow \mathbb{R}^3$. Accordingly, the expected magnetometer measurement at time $t_k$ can be expressed as
\begin{equation}
		\small{
		\label{eq:loc_relation}
		\begin{aligned}
			[\hat{\mathbf{B}}_{\text{meas}}]_{t_k}^i
			&= {\mathbf{R}_{t_k}^i}^{\top} \mathcal{M}(\mathbf{p}_{t_k}^i) + \mathbf{n}_{t_k}^i \\
			&= g\!\left[\mathcal{M}(\cdot), f^i(\mathbf{x}_{t_k})\right] + \mathbf{n}_{t_k}^i,
			\quad i = 1, \dots, N
		\end{aligned}
	}
\end{equation}
where $\mathbf{n}_{t_k}^i$ denotes the measurement noise, and $g(\cdot)$ represents the  transformation that maps the magnetic map readings to the robot body frame. Given the robot state $\mathbf{x}_{t_k}$, the localization problem can be formulated by defining a loss function that quantifies the discrepancy between the measured magnetic value and the map-predicted values:
\begin{equation}
		\small{
		\label{eq:opt_localization}
		S(\mathbf{x}_{t_k}) =
		\sum_{i=1}^{N}
		\left\|
		\mathbf{B}_{t_k}^i -
		g\!\left[\mathcal{M}(\cdot), f^i(\mathbf{x}_{t_k})\right]
		\right\|^2
	}
\end{equation}

\subsection{Magnetic Sensor Intrinsic Calibration Modeling}
\begin{figure}[!t]
	\centering
	\vspace{-1.5em}
	\resizebox{0.90\linewidth}{!}{%
		\begin{minipage}{\linewidth}
			\centering
			\subfloat[Interior Instrumental Imperfection\label{fig:intrumental_dist}]{%
				\includegraphics[width=0.52\linewidth]{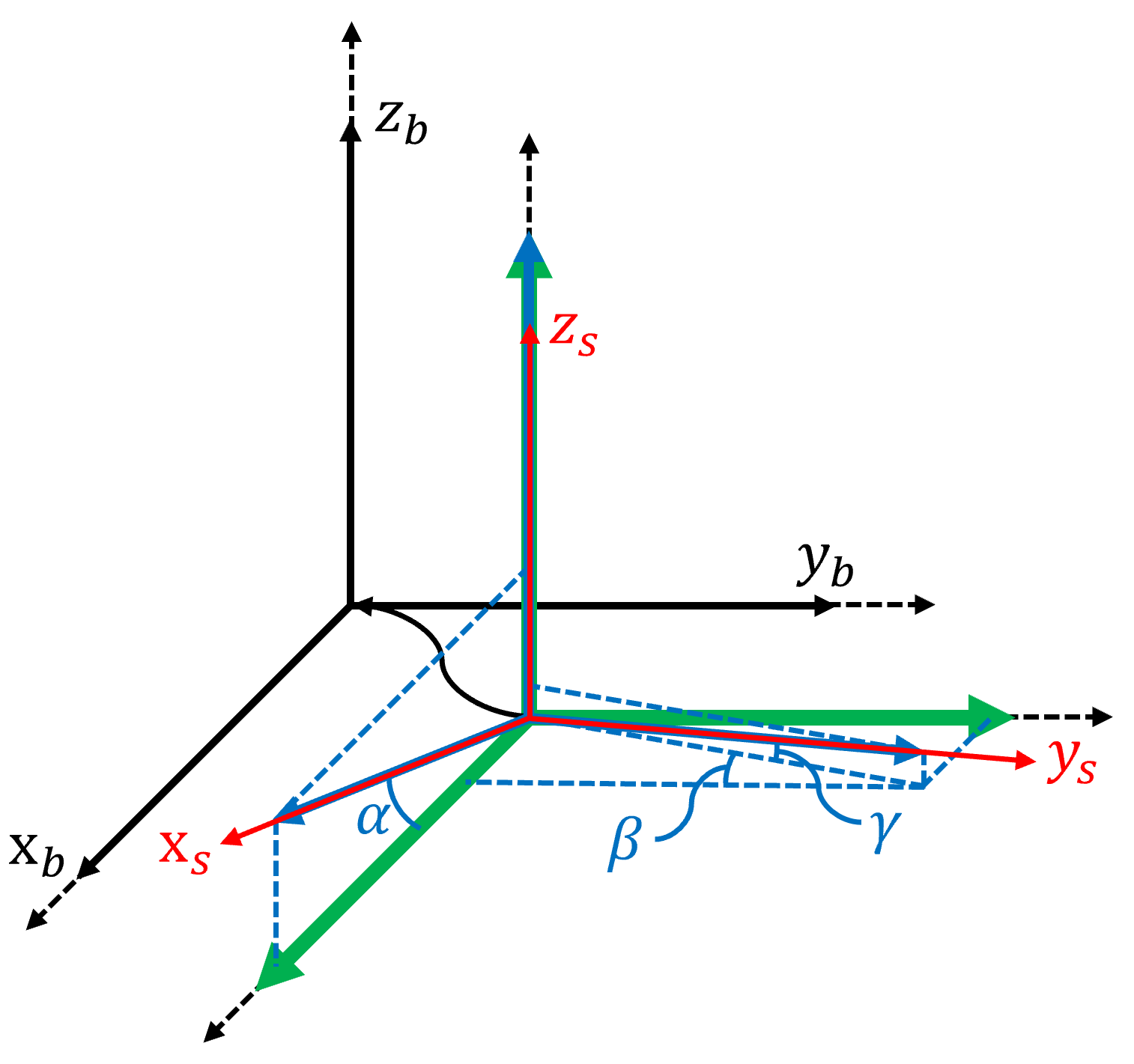}}
			\hfill
			\subfloat[Exterior Platform's Ferromagnetic Materials Distortion\label{fig:ferromagnetic_dist}]{%
				\includegraphics[width=0.475\linewidth]{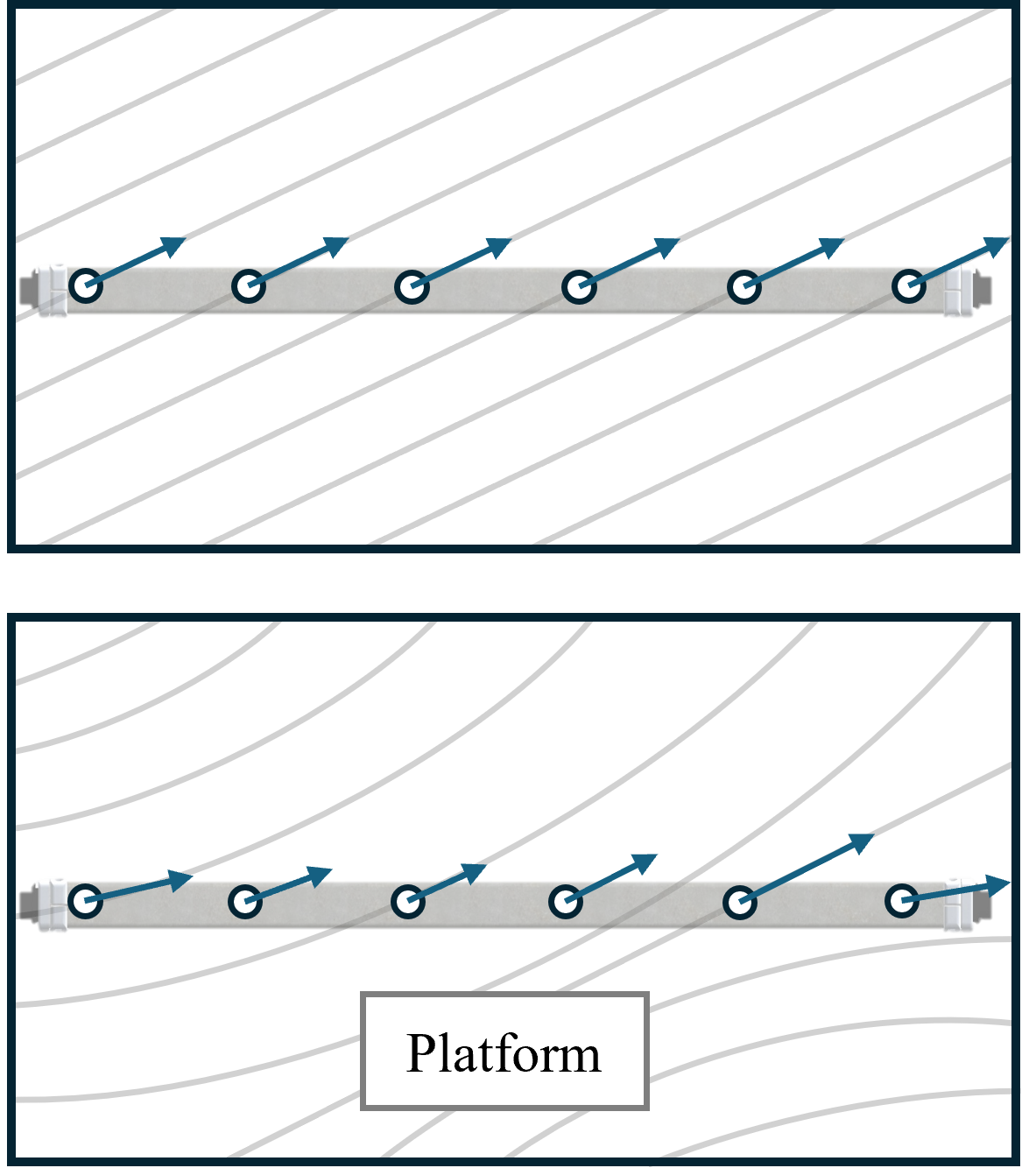}}
		  \end{minipage}
	}
	\caption{The measured values of a magnetic sensor can be influenced by two main factors: (a) interior instrumental imperfection of the magnetometer, and (b) the presence of exterior ferromagnetic materials in the platform.}
	\label{fig:intrinsic_distortion_model}
	\vspace{-1em}
\end{figure}

\begin{figure*}[!t]
	\centering
	\includegraphics[width=0.85\linewidth]{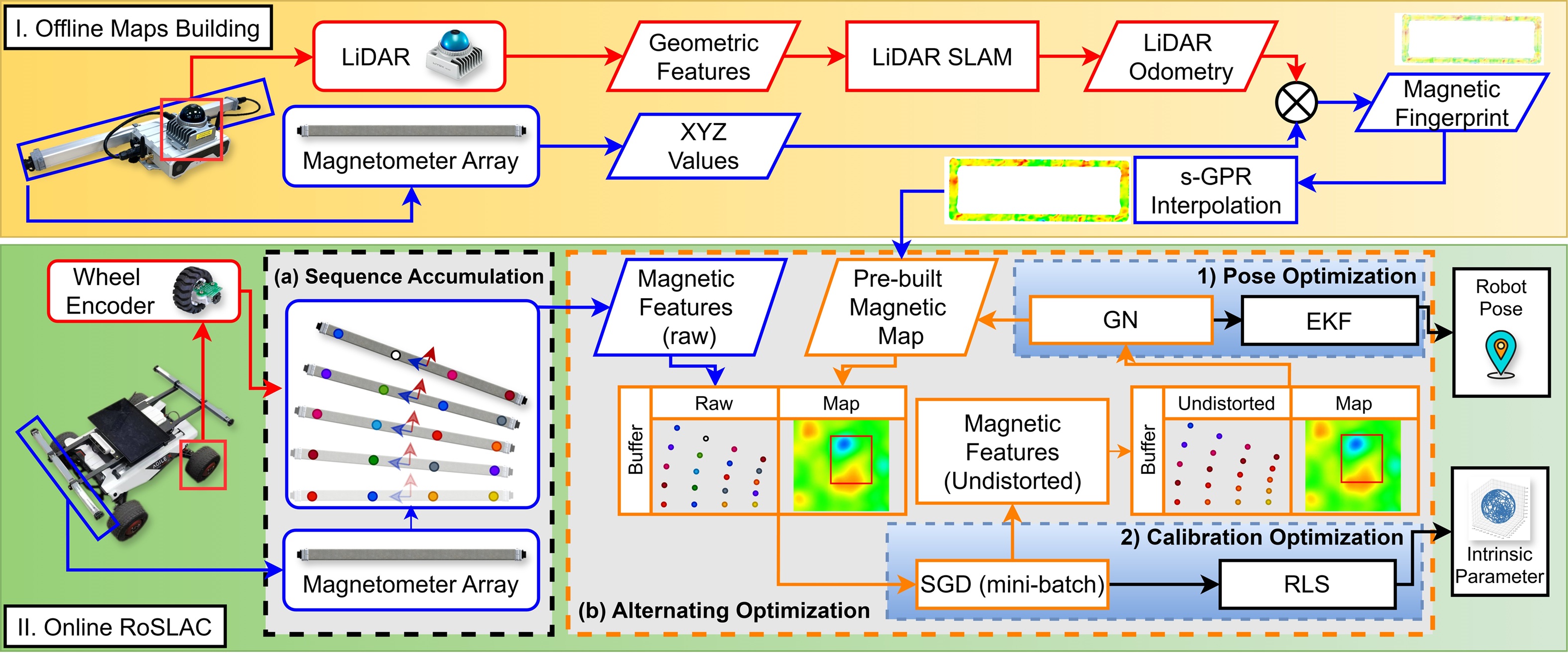}
	\caption{Flowchart of the proposed method. The overall pipeline can be decomposed into two sub-modules: (a) ambient magnetic map pre-building, and (b) the online \textit{RoSLAC} procedure, featuring sequence accumulation and alternating optimization.}
	\label{fig:flowchart}
	\vspace{-1.5em}
\end{figure*}
The previous section implicitly assumed that the magnetometer measurement $\mathbf{B}_{t_k}^i$ is ideal and free of distortion, such that it directly represents AMF expressed in the magnetometer frame, \textit{i.e.},
$\mathbf{B}_{t_k}^i = {[\mathbf{B}_{env}]}_{t_k}^i$.
In practice, however, magnetometer measurements are subject to distortions arising primarily from two sources.

The first source is the magnetometer's intrinsic imperfections, which can cause inconsistent readings across sensors under the same external magnetic field. As shown in~\cite{yangCorrectionMethodThreeAxis2022}, the main error sources are axis-dependent scale factors $\mathbf{S}$, axis non-orthogonality $\mathbf{C}_{NO}$, and constant bias $\mathbf{B}_O$ (Fig.~\ref{fig:intrumental_dist}). Accordingly, the true magnetic field $\tilde{\mathbf{B}}_{t_k}^i$ and the raw measurement $\mathbf{B}_{t_k}^i$ satisfy $\tilde{\mathbf{B}}_{t_k}^i = \mathbf{C}_{NO}^i \mathbf{S}^i \mathbf{B}_{t_k}^i + \mathbf{B}_O^i$. This type of distortion is independent of the surrounding environment and depends solely on the sensor itself. Hence, it can in principle be compensated through magnetometer calibration alone.

The second source of distortion is caused by external ferromagnetic materials on the sensor carrier or platform, which perturb the ambient magnetic field $\mathbf{B}_{env}$ (Fig.~\ref{fig:ferromagnetic_dist}). According to the Tolles-Lawson model~\cite{tollesCompensationAircraftMagnetic1954}, this distortion consists of hard-iron bias $\mathbf{B}_h$ and soft-iron distortion $\mathbf{K}_s$, while the eddy-current effect is neglected in this work~\cite{Li2023Compensation}. Thus, the recovered environmental magnetic field (or equivalently the magnetic map value when a pre-built map is available) is given by ${[\hat{\mathbf{B}}_{meas}]}_{t_k}^i = \mathbf{K}_s^i \tilde{\mathbf{B}}_{t_k}^i + \mathbf{B}_h^i$.

For simplicity, the intrinsic sensor imperfections and the carrier-induced distortions can be combined into a unified affine model:
\begin{equation}
		\small{
		\label{eq:full_carrier_effect}
		{[\hat{\mathbf{B}}_{meas}]}_{t_k}^i
		= \mathbf{C}^i \mathbf{B}_{t_k}^i + \mathbf{b}^i
	}
\end{equation}
where $\mathbf{C}^i = \mathbf{K}_s^i \mathbf{C}_{NO}^i \mathbf{S}^i \in \mathbb{R}^{3 \times 3}$ and
$\mathbf{b}^i = \mathbf{K}_s^i \mathbf{B}_O^i + \mathbf{B}_h^i \in \mathbb{R}^3$
collectively represent the twelve calibration parameters. These parameters are further grouped into a parameter vector
\begin{equation}
	\small{
		\boldsymbol{\theta}^i
		=
		\left[
		{\mathbf{c}_1^i}\;
		{\mathbf{c}_2^i}\;
		{\mathbf{c}_3^i}\;
		b_1^i\;
		b_2^i\;
		b_3^i
		\right]^\top
		=
		\left[
		{\boldsymbol{\theta}^{i,C}}^\top\;
		{\boldsymbol{\theta}^{i,b}}^\top
		\right]^\top
	}
\end{equation}
where $\mathbf{c}_j^i = [c_{j1}^i\; c_{j2}^i\; c_{j3}^i]$ denotes the $j$-th row of the matrix $\mathbf{C}^i$, and $b_j^i$ is the $j$-th element of the bias vector $\mathbf{b}^i$. By rearranging~\eqref{eq:full_carrier_effect}, the magnetometer calibration model can be rewritten in a linear regression form:
\begin{equation}
	\small{
	\label{eq:calib_relation}
	\begin{aligned}
		{[\hat{\mathbf{B}}_{meas}]}_{t_k}^i &=
		\begin{bmatrix}
			{\mathbf{B}_{t_k}^{i}}^{\top} & \mathbf{0}_{1\times3} & \mathbf{0}_{1\times3} & 1 & 0 & 0 \\
			\mathbf{0}_{1\times3} &{\mathbf{B}_{t_k}^{i}}^{\top} & \mathbf{0}_{1\times3} & 0 & 1 & 0 \\
			\mathbf{0}_{1\times3} & \mathbf{0}_{1\times3} & {\mathbf{B}_{t_k}^{i}}^{\top} & 0 & 0 & 1
		\end{bmatrix}
		\boldsymbol{\theta}^i + \boldsymbol{\epsilon}_{t_k} \\
		&= h(\mathbf{B}_{t_k}^{i})\boldsymbol{\theta}^i + \boldsymbol{\epsilon}_{t_k}=\mathbf{H}_{t_k}^i\boldsymbol{\theta}^i+ \boldsymbol{\epsilon}_{t_k}
	\end{aligned}
	}
\end{equation}
where $\boldsymbol{\epsilon}_{t_k}$ denotes the measurement noise. This formulation makes it evident that the overall intrinsic calibration of the magnetometer can be expressed as a linear problem with respect to the calibration parameters. Consequently, the calibration parameters can be estimated via linear optimization by minimizing the following loss function:
\begin{equation}
	\small{
	\label{eq:opt_eq_calib}
	S(\boldsymbol{\theta}^i) = \sum_{k=0}^{T}\left\|h(\mathbf{B}_{t_k}^{i})\boldsymbol{\theta}^i - {[\mathbf{B}_{ref}]}_{t_k}^i\right\|^2
}
\end{equation}

\section{Robust Simultaneous Localization and Calibration (\textit{RoSLAC}) Framework} \label{sec:method}
\subsection{Magnetic Mapping Method}
As discussed in Section~\ref{sec:model}, both localization and calibration rely on an accurate dense magnetic map. In this work, the map is constructed offline using Gaussian Process Regression (GPR) to ensure high accuracy. GPR is a non-parametric, data-driven method that models the magnetic field as a continuous function using correlations captured by a kernel function~\cite{Rasmussen2005}. Let $S=\{(\mathbf{p}_1,\mathbf{B}_1),(\mathbf{p}_2,\mathbf{B}_2),\ldots,(\mathbf{p}_n,\mathbf{B}_n)\}
$ be the training set of magnetic fingerprints, where \(\mathbf{p}\in\mathbb{R}^{n\times 3}\) denotes the 3-D positions and \(\mathbf{B}\in\mathbb{R}^{n\times 3}\) the corresponding three-axis magnetic measurements. GPR is then used to infer the posterior of the latent magnetic field function \(f(\mathbf{p})\) conditioned on \(S\). Following~\cite{lyu2024sgpr}, the relation between \(f(\mathbf{p})\) and the measurement \(\mathbf{B}_i\) at \(\mathbf{p}_i\) is given by:
\begin{equation} 
	\small{
			\begin{split}
				f(\mathbf{p}){\,}{\sim}{\,}\mathcal{GP}{({m(\mathbf{p}), k(\mathbf{p}_j, \mathbf{p}_k))}} \\
				\mathbf{B}_i = f(\mathbf{p}_i) + \epsilon
			\end{split}
		}
\end{equation}
where $m(\mathbf{p})$ denotes the mean function, which is set to the local-scale Earth magnetic field value, and  $\epsilon$ represents zero-mean Gaussian noise with a known variance ${\sigma}_n^2$. Given two input locations $\mathbf{p}_j$ and $\mathbf{p}_k$, the corresponding target magnetic measurements $\mathbf{B}_j$ and $\mathbf{B}_k$ are assumed to be correlated through a covariance (kernel) function, which is expressed as:
\begin{equation}
	\small{
			\operatorname*{cov}(\mathbf{p}_j, \mathbf{p}_k) = k(\mathbf{p}_j, \mathbf{p}_k) + {\sigma}_n^2{\delta}_{jk}
		}
\end{equation}
where $k(\mathbf{p}_j, \mathbf{p}_k)$ denotes the kernel function and $\delta_{jk}$ is the Kronecker delta, which equals $1$ when $p = q$ and $0$ otherwise. The performance of the GPR-based magnetic mapping is fully determined by the collected magnetic training data and the chosen kernel function. In this work, the widely used radial basis function (RBF) kernel is adopted \cite{kok2018scalable}. However, since conventional GPR exhibits high computational complexity, our previous work, s-GPR \cite{lyu2024sgpr}, is employed to accelerate the magnetic map construction process. After obtaining the dense magnetic grid map, magnetic field values at arbitrary query points are computed via bilinear interpolation on the grid, producing continuous spatial estimates. 
\vspace{-1em}
\subsection{Simultaneous Localization and Calibration on Prebuilt Maps}
Traditionally, to estimate all parameters in Eq.~(\ref{eq:opt_eq_calib}), the sensors are rotated in all directions at a fixed position within a homogeneous magnetic field to obtain sufficient observations. However, when sensors are mounted on a heavy or large-scale platform, physically rotating the platform is often impractical, and generating or accessing a sufficiently homogeneous magnetic field can also be difficult.

As an alternative, observability can be achieved by effectively rotating the external magnetic field. This can be realized by placing the platform at multiple known positions within a known, non-homogeneous ambient magnetic field, which can be represented by a pre-built dense grid map as introduced earlier. Although this approach may not provide ground-truth environmental magnetic measurements, it enables unifying all magnetometers to a consistent scale with respect to the pre-built map. A key challenge lies in determining the position of the magnetometer within the map in order to obtain $\hat{\mathbf{B}}_{\mathrm{meas}}$. One feasible solution is to use an auxiliary sensor, such as LiDAR, to perform relocalization and provide the pose $\mathbf{y}_{t_k}^i$ as input~\cite{lyu2025l2mcalibonekeycalibrationmethod}. Nevertheless, relying on an auxiliary LiDAR sensor introduces additional drawbacks such as difficulties in obtaining accurate prior knowledge of temporal synchronization and the extrinsic transformation function $f^i$ in practice. Consequently, it would be more convenient and practical if the robot could directly obtain localization information using the magnetometers themselves, while simultaneously performing calibration.

Thus, we introduce a robust simultaneous localization and calibration method (\textit{RoSLAC}), which is capable of calibrating multiple magnetometers within an array while performing precise localization. To this end, the overall optimization objective is reformulated to jointly incorporate the robot state vector $\mathbf{x}_{t_k}$ and the calibration parameter vector $\boldsymbol{\theta}^i$. By aligning Eq.~(\ref{eq:loc_relation}) with Eq.~(\ref{eq:calib_relation}), the final optimization model can be derived as
\begin{equation}
	\small{
		\label{eq:full_optimization_goal}
		\argmin_{\boldsymbol{\theta}^i}\sum_{k=0}^{T}\argmin_{\mathbf{x}_{t_k}}\sum_{i=1}^{N}
		\left\|h(\mathbf{B}_{t_k}^{i})\boldsymbol{\theta}^i 
		- g\!\left[\mathcal{M}(\cdot), f^i(\mathbf{x}_{t_k})\right]\right\|^2
	}
\end{equation}

To obtain the optimal estimates of $\boldsymbol{\theta}^i$ and $\mathbf{x}_{t_k}$, the overall workflow of the proposed method is illustrated in Fig.~\ref{fig:flowchart}. First, a dense magnetic grid map is constructed offline using Gaussian Process Regression (GPR) and then provided to \textit{RoSLAC} for online state and parameter estimation. However, due to the limited observability of magnetic information and the high dimensionality of the estimation variables, the optimization problem can become ill-conditioned. To address this issue, two strategies are adopted. First, \textit{RoSLAC} accumulates a sequence of measurements online with wheel odometry to enhance observability. Second, an alternating optimization scheme is employed, in which the estimation variables are decoupled and optimized iteratively to reduce the effective state dimension at each step. Finally, the estimated states and parameters are passed to a filtering module to produce the final output. The details of each module are described in the following sections.
\vspace{-2em}
\subsection{Sequence Accumulation}
To enhance local observability, the received magnetic measurements are first accumulated with the assistance of wheel odometry. To incorporate this information, assume that the forward incremental motion provided by the wheel odometry from time $t_{j-1}$ to $t_j$ is given by $(\Delta\mathbf{R}_{t_j}, \Delta\mathbf{p}_{t_j})$, where $j$ denotes a time index preceding the current time $k$ (\textit{i.e.}, $j<k$). The corresponding backward relative pose of frame $t_j$ expressed in the coordinate frame of $t_k$ can then be written as
\begin{equation}
	\small{
		\begin{aligned}
			{}^{t_k}\mathbf{R}_{t_j} &= \prod_{m=k}^{j+1}\Delta\mathbf{R}_{t_m}^{\top}, \\
			{}^{t_k}\mathbf{p}_{t_j} &= -\sum_{m=j+1}^{k}
			\left(\prod_{n=k}^{m+1}\Delta\mathbf{R}_{n}^{\top}\right)
			\Delta\mathbf{R}_{m}^{\top}\,\Delta\mathbf{p}_{m}
		\end{aligned}
	}
\end{equation}
For the special case $j=k$, the relative pose degenerates to ${}^{t_k}\mathbf{R}_{t_j}=\mathbf{I}$ and ${}^{t_k}\mathbf{p}_{t_j}=\mathbf{0}$. Since the data arrive sequentially, the above formulation is undefined for $j>k$.

Given a sliding time window of size $l$, Eq.~(\ref{eq:non_seq_loc}) can be extended to an accumulated pose-sequence representation as
\begin{equation}
	\small{
		\begin{aligned}
			f^i(\mathbf{x}_{t_k}) :\;& \mathbf{x}_{t_k}
			\;\longmapsto\;
			\left\{\left(\mathbf{R}_{t_{k,j}}, \mathbf{p}_{t_{k,j}}\right)\right\}_{j=k-l+1}^{k} \\
			=\;&
			\Big\{
			\big(
			\exp(\boldsymbol{\phi}_{t_k}^{\wedge})\,{}^{t_k}\mathbf{R}_{t_j}\mathbf{R}_m^i,\;
			\exp(\boldsymbol{\phi}_{t_k}^{\wedge})
			\big({}^{t_k}\mathbf{p}_{t_j} + {}^{t_k}\mathbf{R}_{t_j}\mathbf{p}_m^i\big)
			+ \mathbf{p}_{t_k}
			\big)
			\Big\}_{j=k-l+1}^{k}
		\end{aligned}
	}
\end{equation}
The corresponding magnetometer measurements associated with these accumulated poses are expressed as
\begin{equation}
	\small{
		h(\mathbf{B}_{t_k}^i):\;
		\mathbf{B}_{t_k}^i
		\;\longmapsto\;
		\left\{\mathbf{H}_{t_{k,j}}^{i}\right\}_{j=k-l+1}^{k}
	}
\end{equation}
\vspace{-2em}
\subsection{Alternating Optimization}
Even with the accumulated data sequence, directly solving the above optimization problem remains challenging due to the high dimensionality of the combined state and parameter vectors to be estimated. Fortunately, as shown in Section~\ref{sec:model}, the optimization can be decoupled into the state variables and the calibration parameters, enabling the use of an alternating optimization strategy. In the following, we first focus on the optimization problem at the current time step $t_k$. The two sub-optimization problems are:
\subsubsection{Calibration Parameter Optimization with Stochastic Gradient Descent (SGD)} Given the robot pose state $\mathbf{x}_{t_k}$, the cost function depends solely on the calibration parameter vector. Specifically, the optimization problem can be written as
\begin{equation}
	\small{
		\begin{aligned}
			\boldsymbol{\theta}^{i,\ast}_{t_k}
			&= \argmin_{\boldsymbol{\theta}^{i}_{t_k}}
			\left\| h(\mathbf{B}_{t_k}^{i}) \boldsymbol{\theta}^{i}_{t_k}
			- g\!\left[\mathcal{M}(\cdot), f^i(\mathbf{x}_{t_k})\right] \right\|^2
			+ \lambda \left\| \boldsymbol{\theta}^{i,C}_{t_k} \right\|^2
			\label{eq:opt_goal1}
		\end{aligned}
	}
\end{equation}

Ideally, as discussed in Section~\ref{sec:model}, the calibration model is linear and admits a closed-form solution corresponding to the global minimum at $\boldsymbol{\theta}^{i,*}$. However, due to the limited amount of information available at each time step, directly solving the least-squares problem may suffer from ill-conditioned data, leading to numerical instability caused by the inversion of the normal matrix. Similarly, the Hessian inversion required by Gauss--Newton methods can also be unstable under such conditions. To avoid explicit matrix inversion, we instead adopt a gradient descent (GD) strategy, since GD relies only on first-order gradient information and performs parameter updates through iterative descent steps. Moreover, since measurements arrive sequentially, only data up to time $t_k$, \textit{i.e.}, $\mathbf{B}_{0:k}^i$, are available at each step. This naturally motivates the use of a mini-batch stochastic gradient descent (SGD) scheme.

The core idea of mini-batch SGD is to approximate the full gradient computed over the entire dataset by a local gradient evaluated on a batch of samples. To further stabilize the estimation, an $\ell_2$ regularization term is incorporated. Given the objective in Eq.~(\ref{eq:opt_goal1}), the parameter update using the approximate gradient is given by
\begin{equation}
	\small{
	\boldsymbol{\theta}^{i}_{t_k, m+1}
	=
	\boldsymbol{\theta}^{i}_{t_k, m}
	- \eta \, \nabla \mathbf{r}(\boldsymbol{\theta}^{i}_{t_k, m})
}
\end{equation}
where $\eta$ denotes the learning rate and $m$ is the iteration index at time $t_k$. The gradient is computed as
\begin{equation}
	\small{
		\begin{aligned}
			\nabla \mathbf{r}(\boldsymbol{\theta}^{i}_{t_k, m})&=\\
			\sum_{j=k-l+1}^{k}&
			{\mathbf{H}_{t_{k,j}}^{i}}^{\top}
			\Big(
			\mathbf{H}_{t_{k,j}}^{i} \boldsymbol{\theta}^{i}_{t_k, m}
			-
			g\!\left[\mathcal{M}(\cdot),
			\left(\mathbf{R}_{t_{k,j}}, \mathbf{p}_{t_{k,j}}\right)\right]
			\Big)
			+
			\begin{bmatrix}
				\lambda \boldsymbol{\theta}^{i,C}_{t_k} \\
				\mathbf{0}^{3 \times 1}
			\end{bmatrix}
		\end{aligned}
	}
\end{equation}
Since the calibration parameters associated with different magnetometers are independent, the optimal solution $\boldsymbol{\theta}^{i,*}$ is updated separately for each sensor.

\subsubsection{Robot State Optimization with Gauss-Newton}
After obtaining an updated estimate of the calibration parameters in the previous step, the alternating optimization framework allows the robot state to be refined using the calibrated measurements. Focusing on the state at time $t_k$ only, Eq.~(\ref{eq:full_optimization_goal}) can be rewritten as a function of $\mathbf{x}_{t_k}$:
\begin{equation}
		\small{
		\begin{aligned}
			\mathbf{x}_{t_k}^{*}
			&= \argmin_{\mathbf{x}_{t_k}}
			\sum_{i=1}^{N}
			\left\|
			h(\mathbf{B}_{t_k}^{i}) \boldsymbol{\theta}^{i,*}_{t_k}
			-
			g\!\left[\mathcal{M}(\cdot), f^i(\mathbf{x}_{t_k})\right]
			\right\|^2
			\label{eq:opt_goal2}
		\end{aligned}
	}
\end{equation}
In the state estimation problem, the robot state satisfies $\mathbf{x}_{t_k} \in \mathbb{R}^{6}$. Given a single magnetometer measurement $\mathbf{B}_{t_k}^i \in \mathbb{R}^{3}$, the system is underdetermined; however, a sequence of measurements collected by a magnetometer array provides sufficient constraints to render the problem well-posed. The main challenge lies in the inherent nonlinearity of the state estimation problem. In this work, we adopt the classical Gauss--Newton method, which offers a favorable trade-off between estimation accuracy and computational efficiency. Let the perturbation vector be defined as $\Delta\mathbf{x} = \left[\Delta\mathbf{p}^\top\;\delta\boldsymbol{\phi}^\top\right],\ \Delta\mathbf{x} \in \mathbb{R}^{6}$,
where $\Delta\mathbf{p}$ and $\delta\boldsymbol{\phi}$ denote the translational and rotational increments, respectively.
To express the state update on the underlying manifold in a compact form, we adopt the $\boxplus$ operator following~\cite{hertzberg2013integrating}, yielding
\begin{equation}
	\small{
		\mathbf{r}^i(\mathbf{x}_{t_k} \boxplus \Delta\mathbf{x})
		=
		h(\mathbf{B}_{t_k}^{i}) \boldsymbol{\theta}^{i,*}_{t_k}
		-
		\left(
		\mathbf{R}_{t_k}^i \exp\!\left(\delta\boldsymbol{\phi}\right)
		\right)^{\top}
		\mathcal{M}\!\left(\mathbf{p}_{t_k}^i + \Delta\mathbf{p}\right)
	}
\end{equation}

To solve the Gauss-Newton problem, the residual function $\mathbf{r}^i$ is linearized around the current state using a first-order Taylor expansion $\mathbf{r}^i(\mathbf{x}_{t_k} \boxplus\Delta\mathbf{x})\approx\mathbf{r}^i(\mathbf{x}_{t_k})+\mathbf{J}^i \Delta\mathbf{x}$.
Here, $\mathbf{J}^i$ denotes the Jacobian matrix of the $i$-th magnetometer observation with respect to the state perturbation which is given by
\begin{equation}
	\small{
		\begin{aligned}
			\mathbf{J}^i
			&=
			-\sum_{j=k-l+1}^{k}
			\left[
			{\mathbf{R}_{t_{k,j}}^{i}}^{\top}
			\nabla \mathcal{M}(\mathbf{p}_{t_{k,j}}^{i})
			\;\middle|\;
			\big[
			{\mathbf{R}_{t_{k,j}}^{i}}^{\top}
			\mathcal{M}(\mathbf{p}_{t_{k,j}}^{i})
			\big]_{\times}
			\right]
		\end{aligned}
	}
\end{equation}
where $\nabla \mathcal{M}(\mathbf{p}_{t_{k,j}}^{i})$ denotes the spatial gradient of the magnetic grid map evaluated at the current position, and $[\cdot]_{\times}$ represents the skew-symmetric matrix operator.
The Gauss-Newton state update is then obtained as $\mathbf{x}_{t_k, m+1} = \mathbf{x}_{t_k, m}\boxplus\Delta\mathbf{x}$
with the perturbation $\Delta\mathbf{x}$ computed by
\begin{equation}
	\small{
		\Delta\mathbf{x}
		=
		-
		\left(
		\sum_{i=1}^{N}
		{\mathbf{J}^i}^{\top} \mathbf{J}^i
		\right)^{-1}
		\sum_{i=1}^{N}
		{\mathbf{J}^i}^{\top} \mathbf{r}^i
	}
\end{equation}

Within the alternating optimization framework, the calibration parameter optimization and the robot state optimization are executed iteratively until convergence of the overall system is achieved.
\vspace{-1em}
\subsection{Post-iteration Filter} 
In the previous section, the alternating optimization strategy considers only the optimal solution at the current time step $t_k$. However, as indicated by Eq.~(\ref{eq:full_optimization_goal}), the overall optimization objective for $\boldsymbol{\theta}^i$ is inherently a batch problem defined over the entire time horizon $k \in [0, T]$. Therefore, after obtaining the optimal state estimate $\mathbf{x}_{t_k}^*$ at each time step, the outer summation over time must be taken into account to recover the globally optimal calibration parameters $\boldsymbol{\theta}^{i,*}$:
\begin{equation}
	\small{
		\begin{aligned}
			\label{eq:final_theta_optimization_goal}
			\boldsymbol{\theta}^{i,*}
			&= \argmin_{\boldsymbol{\theta}^i}
			\sum_{k=0}^{T}
			\left\|
			h(\mathbf{B}_{t_k}^{i}) \boldsymbol{\theta}^i
			-
			g\!\left[\mathcal{M}(\cdot), f^i(\mathbf{x}_{t_k}^*)\right]
			\right\|^2
		\end{aligned}
	}
\end{equation}
Directly recomputing this batch summation at every time step is computationally expensive. To address this issue, we adopt a recursive least squares (RLS) formulation. For notational simplicity, we define $\mathbf{r}^i = \mathbf{r}(\boldsymbol{\theta}^i)$, $h_k^i = h(\mathbf{B}_{t_k}^{i})$, and $g_k^i = g\!\left[\mathcal{M}(\cdot), f^i(\mathbf{x}_{t_k}^*)\right]$. The optimal calibration parameters $\boldsymbol{\theta}_T^{i,*}$ after time $T$ satisfy the normal equation
$\nabla_{\boldsymbol{\theta}^i}\!\left[\sum_{k=0}^{T} \|\mathbf{r}^i\|^2\right]=\mathbf{0}$.
Let
$\mathbf{P}_T = \sum_{k=0}^{T} {h_k^i}^{\top} h_k^i$
and
$\mathbf{b}_T = \sum_{k=0}^{T} {h_k^i}^{\top} g_k^i$.
Then the closed-form solution can be written as
\begin{equation}
	\small{
		\label{eq:rsl_T}
		\boldsymbol{\theta}_T^{i,*}
		=
		\mathbf{P}_T^{-1} \mathbf{b}_T
	}
\end{equation}
It follows directly that
\begin{equation}
	\small{
		\label{eq:rsl_T_1}
		\begin{aligned}
			\mathbf{P}_T &= \mathbf{P}_{T-1} + {h_T^i}^{\top} h_T^i \\
			\mathbf{b}_T &= \mathbf{b}_{T-1} + {h_T^i}^{\top} g_T^i
		\end{aligned}
	}
\end{equation}
Substituting Eq.~(\ref{eq:rsl_T_1}) into Eq.~(\ref{eq:rsl_T}) yields the recursive update
\begin{equation}
	\small{
		\begin{aligned}
			\boldsymbol{\theta}_T^{i,*}
			&=
			\boldsymbol{\theta}_{T-1}^{i,*}
			+
			\underbrace{
				\frac{\mathbf{P}_{T-1}^{-1} {h_T^i}^{\top}}
				{\mathbf{I} + h_T^i \mathbf{P}_{T-1}^{-1} {h_T^i}^{\top}}
			}_{\text{gain}}
			\left(
			g_T^i - h_T^i \boldsymbol{\theta}_{T-1}^{i,*}
			\right)
		\end{aligned}
	}
\end{equation}
This formulation shows that the optimal calibration parameters at time $T$ depend only on the previous estimate $\boldsymbol{\theta}_{T-1}^{i,*}$, the covariance matrix $\mathbf{P}_{T-1}$, and the current measurements $h_T^i$ and $g_T^i$. Since $\boldsymbol{\theta}^{i,*}$ is assumed to be time-invariant during calibration, the forgetting factor in the RLS update is set to $\lambda = 1$.

\section{ Experiments and Analysis} \label{sec:exp}
\begin{figure*}[!t]
	\centering
	\resizebox{0.85\textwidth}{!}{
		\begin{minipage}{\textwidth}
			\centering
			\subfloat[Pre-built magnetic map and localization results. The color encodes the magnetic field intensity ($\ell_2$-norm). The estimated trajectories and the LiDAR-based ground-truth trajectories are shown in red and blue, respectively. \label{fig:path_result}]{%
				\includegraphics[width=0.74\linewidth]{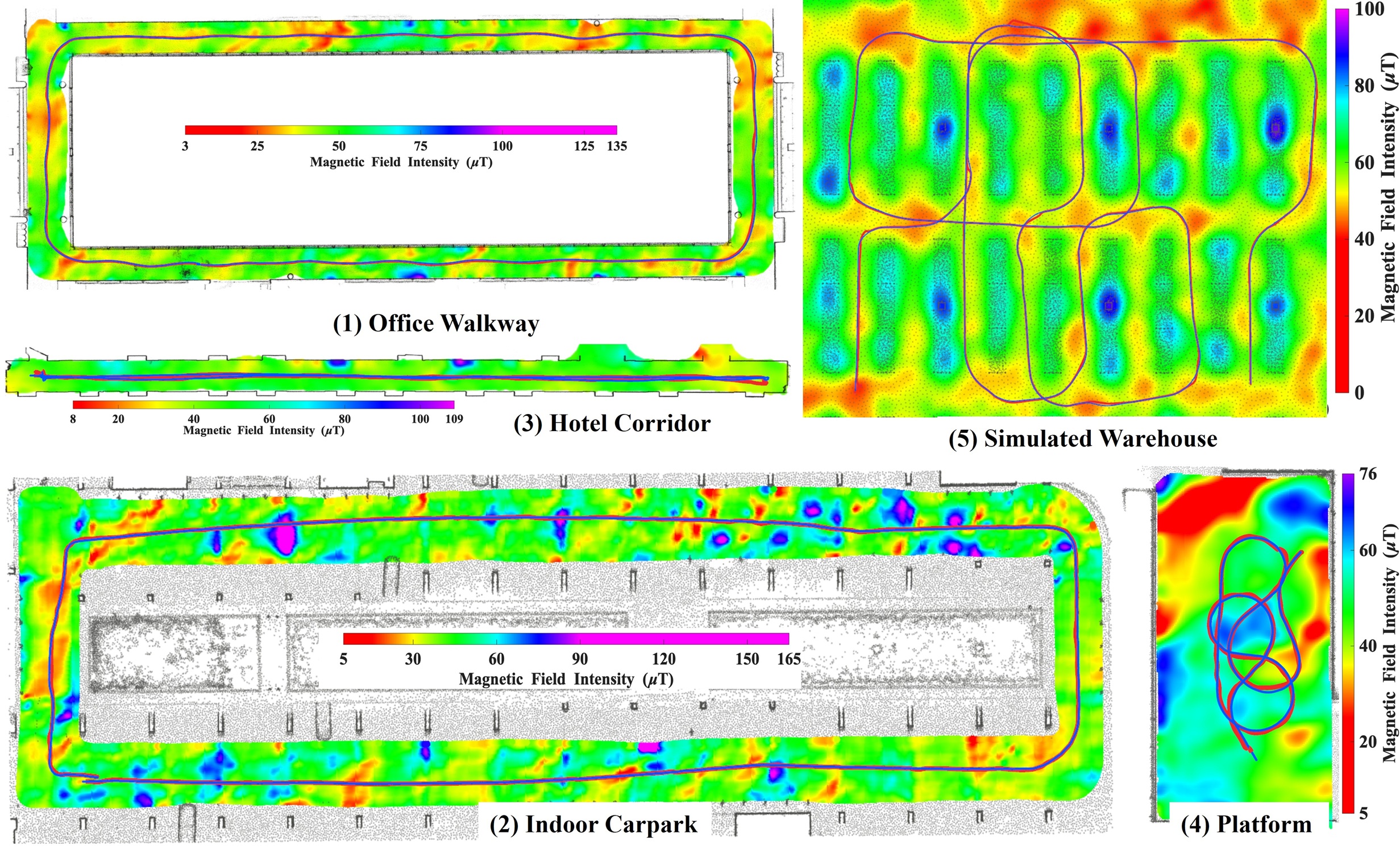}}
			\hfill
			\subfloat[The scenario in which magnetic and LiDAR data is collected.\label{fig:scene}]{%
				\includegraphics[width=0.25\linewidth]{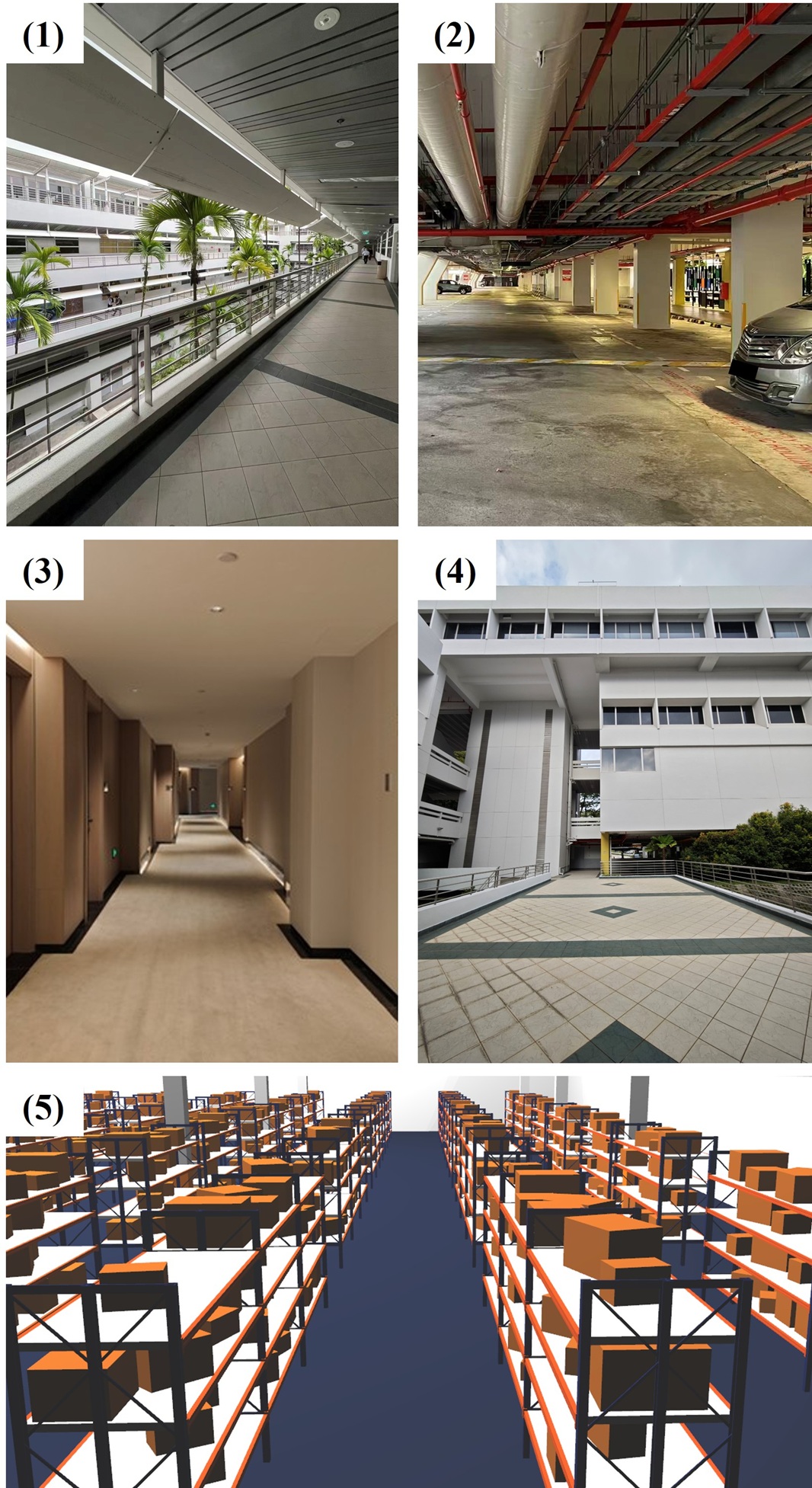}}
		\end{minipage}
	}
	\caption{Five different experimental environments. Environments (1)–(4) correspond to real-world scenarios, while (5) represents a simulated warehouse environment constructed in the Gazebo simulator.}
	\label{fig:test_env}
\end{figure*}
\begin{figure}[!t]
	\centering
	\vspace{-1.5em}
	\resizebox{0.85\linewidth}{!}{%
		\begin{minipage}{\linewidth}
			\centering
			\subfloat[Real Scout Mini AMR\label{fig:real_scout}]{%
				\includegraphics[width=0.50\linewidth]{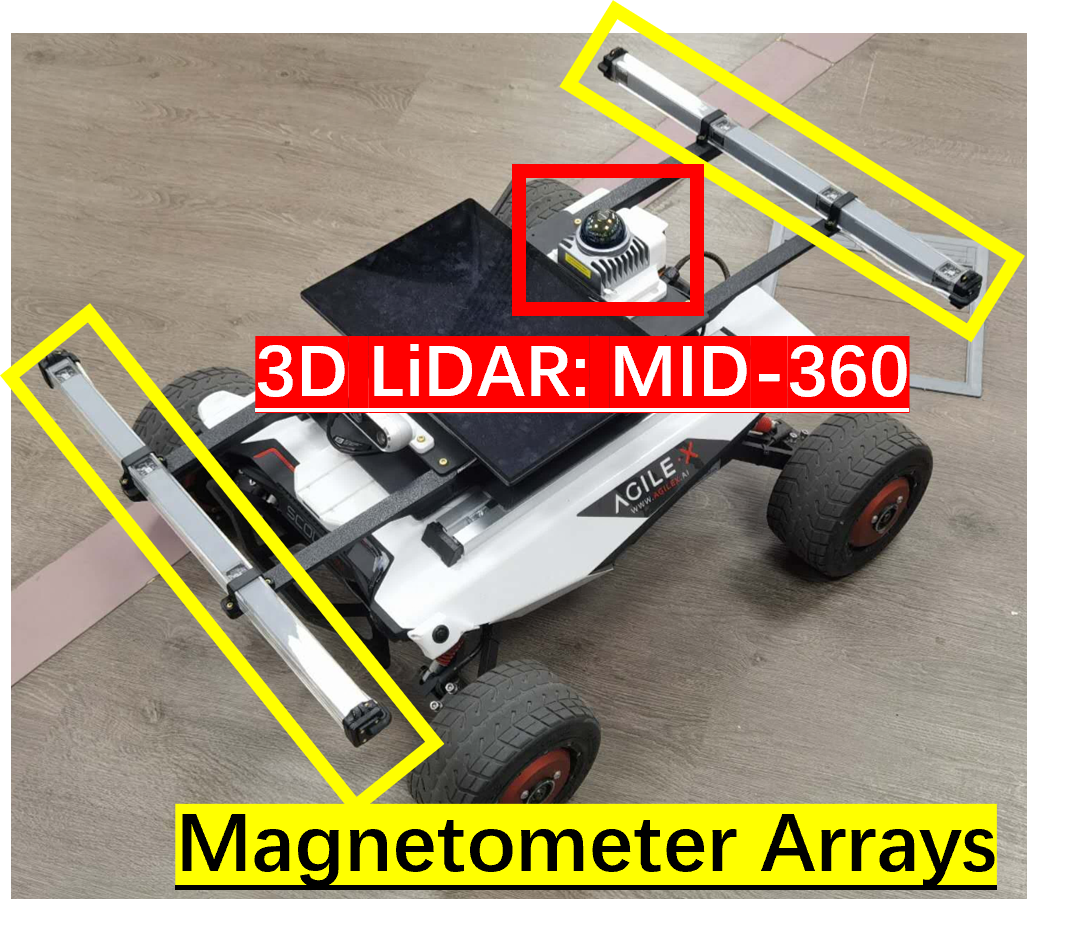}}
			\hfill
			\subfloat[Simulated Scout Mini AMR\label{fig:simu_scout}]{%
				\includegraphics[width=0.48\linewidth]{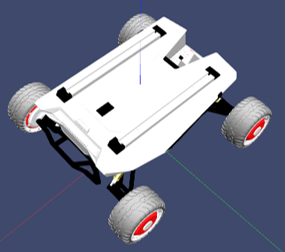}}
	 	 \end{minipage}
	}
	\caption{The Scout Mini AMR is employed in both real-world and simulated environments to collect experimental data.}
	\label{fig:amr_platform}
	\vspace{-1em}
\end{figure}
\subsection{Evaluation Overview}
This section evaluates the proposed \textit{RoSLAC} method in both simulation and real-world environments to verify its effectiveness and robustness. All experiments are conducted on the same desktop computer with an Intel i7 CPU at 2.3\,GHz and 64\,GB RAM. In simulation, an AMR with ground-truth odometry and intrinsic parameters is constructed using the same robot model as in the real-world experiments.

The experimental settings are summarized as follows.
\begin{itemize}
	\item \textbf{High-fidelity Simulations:}  
	Gazebo\footnote{\url{https://gazebosim.org/home}} is used to simulate an industrial warehouse environment. As shown in Fig.~\ref{fig:scene}-(5), the warehouse covers $45\,\mathrm{m} \times 40\,\mathrm{m}$ and includes repetitive storage racks, reinforced concrete pillars, and conveyor belts. A simulated Scout Mini robot (Fig.~\ref{fig:simu_scout}) is equipped with eight idealized magnetic sensors, with four mounted at the front and four at the rear. Zero-mean Gaussian noise with standard deviation $\sigma = 0.2$ is added to the magnetic measurements. Ground-truth poses are obtained directly from Gazebo joint states.
	
	\item \textbf{Real-world Experiments:}  
	A real Scout Mini AMR (Fig.~\ref{fig:real_scout}) is used to collect ambient magnetic field data in four representative environments shown in Fig.~\ref{fig:scene}(1-4): a semi-indoor office walkway ($54\,\mathrm{m} \times 19\,\mathrm{m}$), an indoor carpark ($126\,\mathrm{m} \times 35\,\mathrm{m}$), an indoor hotel corridor ($40\,\mathrm{m} \times 1.5\,\mathrm{m}$), and an outdoor platform ($15\,\mathrm{m} \times 7.5\,\mathrm{m}$). As in simulation, eight RM3100 magnetometers are mounted on the chassis, and a Livox MID-360 LiDAR is installed on top. All sensors are extrinsically calibrated before data collection. Magnetic training points are generated by relocalization with CTE-MLO~\cite{shen2025ctemlo} on a Leica total-station-based prebuilt point cloud map.
\end{itemize}
\vspace{-1em}
\setlength{\arrayrulewidth}{0.2mm}
\subsection{Evaluation Protocol}
\subsubsection{Comparison Baseline}
The proposed method is quantitatively compared with state-of-the-art approaches in terms of localization and calibration accuracy. For localization, each method is evaluated using two types of magnetic inputs: \textit{(a)} raw magnetometer measurements and \textit{(b)} pre-calibrated measurements obtained using~\cite{lyu2025l2mcalibonekeycalibrationmethod}. Four localization methods are considered: \textbf{\textit{PF}}~\cite{shi2022PDR}, a pure localization method based on a particle filter; \textbf{\textit{SO}}~\cite{shenIDFMFLInfrastructurefreeDriftfree2024a}, a pure localization method based on stochastic optimization; \textbf{\textit{RBPF}}~\cite{sieblerMagneticFieldbasedIndoor2023}, a joint localization and calibration approach based on a Rao--Blackwellized particle filter; and \textbf{\textit{RoSLAC}}, the proposed method that performs simultaneous localization and calibration via alternating optimization. For fairness, methods \textit{(a)}--\textit{(c)} use the same number of particles, $n=3000$, as recommended in~\cite{sieblerMagneticFieldbasedIndoor2023}. For calibration, only methods supporting online calibration are evaluated using raw magnetic measurements, namely \textbf{\textit{RBPF}} and \textbf{\textit{RoSLAC}}.

\subsubsection{Evaluation Metrics}
The compared methods are evaluated using localization and calibration accuracy.
\textbf{a) Localization Accuracy:}
In simulation, ground truth is obtained from Gazebo. In real-world experiments, LiDAR odometry relocalized on a Leica total-station-based point cloud map~\cite{shen2025ctemlo} is used as ground truth. Localization performance is measured by the Absolute Trajectory Error (ATE) between the estimated trajectory \(\{\mathbf{T}_{e_i}\}_{i=1}^N\) and the ground-truth trajectory \(\{\mathbf{T}_{g_i}\}_{i=1}^N\). After rigid alignment by \(\mathbf{S}\) and $\mathbf{F}_i = \mathbf{T}_{g,i}^{-1}\left(\mathbf{S}\mathbf{T}_{e,i}\right)$,
the ATE can be computed as
\begin{equation}
	\small{
		\textit{ATE} = \sqrt{\frac{1}{N}\sum_{i=1}^N \left\| \textit{trans}(\mathbf{F}_i) \right\|^2 }
	}
\end{equation}

\textbf{b) Calibration Accuracy:}
In simulation, the predefined calibration parameters are treated as ground truth. In real-world experiments, the offline calibration results from~\cite{lyu2025l2mcalibonekeycalibrationmethod} are used as reference. Let \(\boldsymbol{\theta}_e^*\) and \(\boldsymbol{\theta}_g\) denote the estimated and ground-truth calibration parameters, respectively. The calibration error is defined with $\ell_2$-norm as
\begin{equation}
	\small{
		e_{\boldsymbol{\theta}} = \left\| \boldsymbol{\theta}_e^* - \boldsymbol{\theta}_g \right\|_2
	}
\end{equation}

\subsection{Accuracy Evaluation}
\subsubsection{Localization Accuracy}
We first evaluate localization performance by comparing the ATE of the proposed method with the baselines. As reported in Table~\ref{tab:exp1}, the proposed method achieves the best performance in most scenarios, with an average ATE of about $10\mathrm{cm}$, approaching the accuracy of the LiDAR relocalization reference. When using raw magnetic measurements, it reduces ATE by more than $61\%$ compared with PF and SO. This performance gain arises because PF and SO are purely localization-based methods that fully trust magnetic measurements and do not account for measurement distortions, showing the importance of online calibration under measurement distortion. Even with pre-calibrated measurements, it still improves localization accuracy by over $35\%$ relative to PF and SO. This improvement can be attributed to temporal sequence accumulation, which provides richer observations for localization and enhances the model’s ability to converge to a globally optimal solution.

For RBPF and \textit{RoSLAC}, pre-calibration has little effect on localization accuracy, since both methods jointly perform online calibration and localization. This behavior is expected, as both RBPF and \textit{RoSLAC} jointly perform online calibration and localization. In the \textit{Office} scenario, \textit{RoSLAC} achieves an ATE approximately $1\mathrm{cm}$ lower than RBPF, although this difference is not statistically significant due to the narrow office walkway width. As shown in Fig.~\ref{fig:path_result}, the estimated trajectories closely match the LiDAR-based ground truth in all scenarios, further confirming the accuracy of the proposed method.

\subsubsection{Calibration Accuracy}
\begin{figure}[t]
	\centering
	\includegraphics[width=0.85\linewidth]{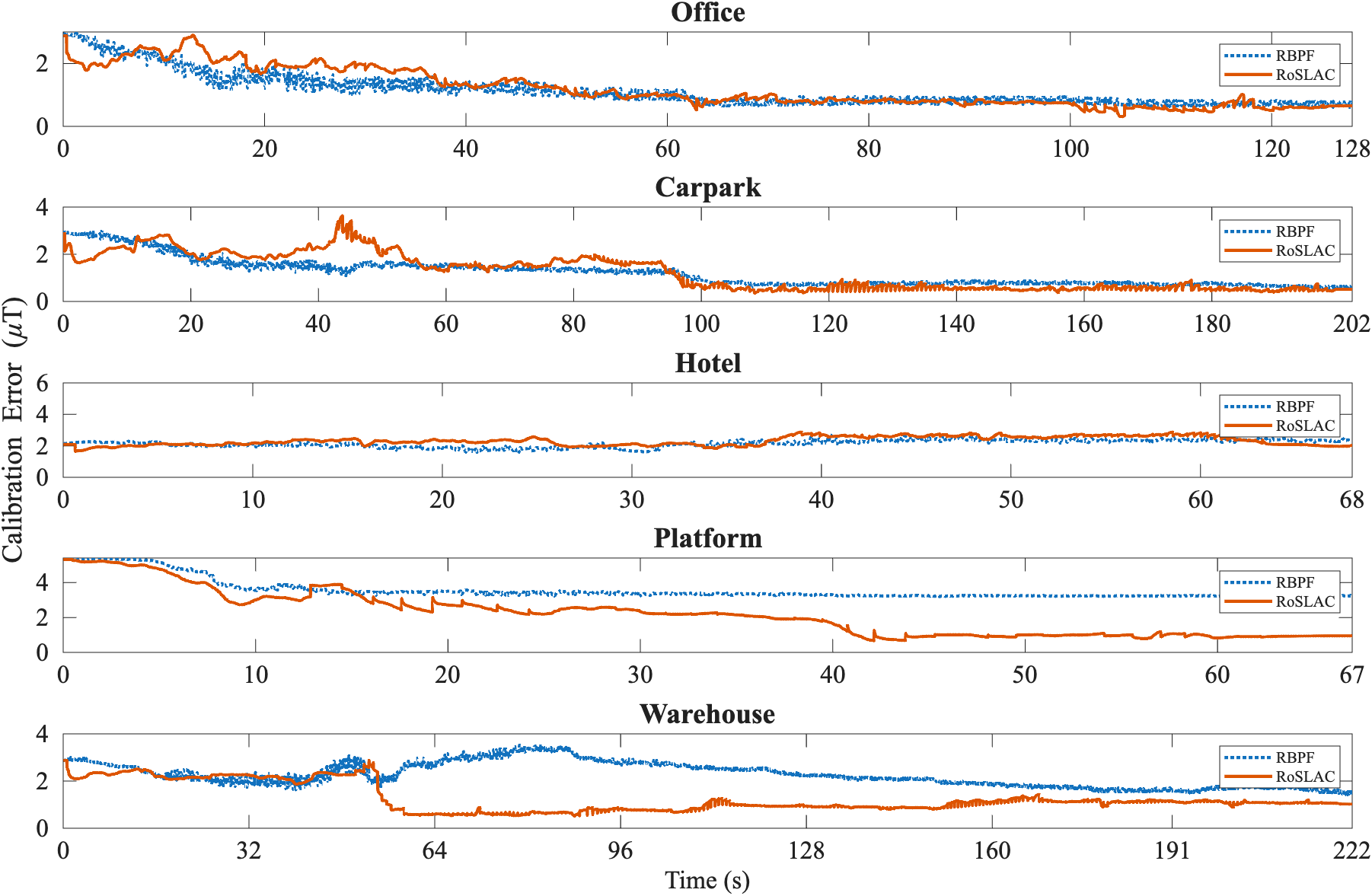}
	\caption{Visualization of the convergence of the magnetometer calibration parameters, measured as the $\ell_2-norm$ deviation from the reference values, under different scenarios.}
	\label{fig:exp4}
	\vspace{-1em}
\end{figure}

\begin{table}[]
	\belowrulesep=0pt
	\aboverulesep=0pt
	\setlength{\tabcolsep}{3.5pt}
	\centering
	\caption{The localization accuracy comparison in Absolute Trajectory Error (ATE, meter). The best result is highlighted in \textbf{BOLD}}
	\label{tab:exp1}
	\vspace{-1em}
	\renewcommand{\arraystretch}{1.2}
	\resizebox{0.95\linewidth}{!}{%
		\begin{threeparttable}
			\begin{tabular}{l||cccccccc}
				\hline\hline
				& \multicolumn{8}{c}{Localization ATE (m)} \\ \cline{2-9} 
				\multicolumn{1}{c||}{\multirow[b]{2}{*}{Scenario}} & \multicolumn{2}{c}{\textbf{RS}} & \multicolumn{2}{c}{RP} & \multicolumn{2}{c}{SO} & \multicolumn{2}{c}{PF} \\ \cmidrule(lr){2-3}\cmidrule(lr){4-5}\cmidrule(lr){6-7}\cmidrule(lr){8-9} 
				\multicolumn{1}{c||}{} & raw & p-cal & raw & p-cal & raw & p-cal & raw & p-cal \\ \hline
				Office & 0.114 & 0.114 & \textbf{0.103} & \textbf{0.107} & 0.167 & 0.130 & 0.153 & 0.121 \\
				Carpark & \textbf{0.103} & \textbf{0.104} & 0.127 & 0.132 & 0.414 & 0.161 & 0.308 & 0.146 \\
				Hotel & \textbf{0.117} & 0.118 & 0.208 & 0.195 & 0.448 & 0.133 & 0.163 & \textbf{0.112} \\
				Platform & \textbf{0.093} & \textbf{0.083} & 0.100 & 0.112 & 0.886 & 0.199 & 0.260 & 0.194 \\
				Warehouse & \textbf{0.087} & \textbf{0.065} & 0.342 & 0.489 & 0.498 & 0.142 & 0.453 & 0.150 \\ \hline
				\multicolumn{1}{r||}{Average} & \textbf{0.103} & \textbf{0.097} & 0.176 & 0.207 & 0.483 & 0.153 & 0.267 & 0.145 \\ \hline\hline
			\end{tabular}
			
			\begin{tablenotes}
				\footnotesize
				\item[1] RS, RP denote proposed \textit{RoSLAC} and RBPF, raw/p-cal denote un-/pre-calibrated.
			\end{tablenotes}
		\end{threeparttable}
	}
\end{table}
Calibration accuracy is further evaluated by comparing the estimated calibration parameter vector with the offline calibration results. As only RBPF and the proposed \textit{RoSLAC} support online calibration, the comparison is limited to these two methods. The calibration trajectories are shown in Fig.~\ref{fig:exp4}, and the quantitative results are reported in Table~\ref{tab:exp3}. As illustrated in Fig.~\ref{fig:exp4}, the proposed \textit{RoSLAC} drives the calibration error to a low converged value. In the \textit{Office} and \textit{Carpark} scenarios, small fluctuations are observed, mainly due to SGD-based optimization under ill-conditioned estimation. Nevertheless, once sufficiently informative observations are available, the calibration parameters quickly converge to near-optimal values. By comparison, RBPF exhibits smoother convergence in these cases because of its probabilistic filtering mechanism. However, in the \textit{Platform} and \textit{Warehouse} scenarios, RBPF tends to become overconfident and get trapped in local optima, as further discussed in the robustness analysis.

The final calibration accuracy is summarized in Table~\ref{tab:exp3}. Consistent with the localization results, the proposed method achieves high calibration accuracy in most scenarios. Relative to the offline calibration baseline, the average calibration error is reduced to about $0.5\mu\mathrm{T}$ in the \textit{Carpark} scenario. Overall, the proposed method remains consistent across environments, except in the \textit{Hotel} scenario, where the error is larger. This is likely caused by the simple calibration trajectory, which limits observability even near the end of the process. Notably, this reduced calibration accuracy does not significantly degrade localization performance. One possible reason is that the joint calibration-localization objective may admit multiple local minima, allowing accurate local pose estimation under different calibration parameters. However, such locally optimal solutions may not generalize to other locations, potentially reducing long-term robustness. Therefore, obtaining a globally optimal calibration solution remains important.

\begin{table}[t]
	\centering
	\caption{The final calibration parameter Error ($\ell_2-norm$, $\mu$T). The best result is highlighted in \textbf{BOLD}}
	\label{tab:exp3}
	\vspace{-1em}
	\begin{adjustbox}{width=\linewidth}
		\begin{tabular}{l||ccccc||c}
			\hline\hline
			\multirow[b]{2}{*}{Method} & \multicolumn{6}{c}{Calibration Error ($\mu$T)} \\ \cline{2-7} 
			& Office & Carpark & Hotel & Platform & Warehouse & Average \\ \hline
			\textbf{RoSLAC} & \textbf{0.659} & \textbf{0.523} & \textbf{2.055} & \textbf{0.950} & \textbf{1.017} & \textbf{1.041} \\
			RBPF & 0.663 & 1.001 & 3.181 & 3.653 & 4.138 & 2.527 \\ \hline\hline
		\end{tabular}
	\end{adjustbox}
	\vspace{-2em}
\end{table}

\vspace{-0.9em}
\subsection{Robustness Evaluation}
\subsubsection{Localization Robustness}
\begin{figure*}[!t]
	\centering
	\includegraphics[width=0.85\linewidth]{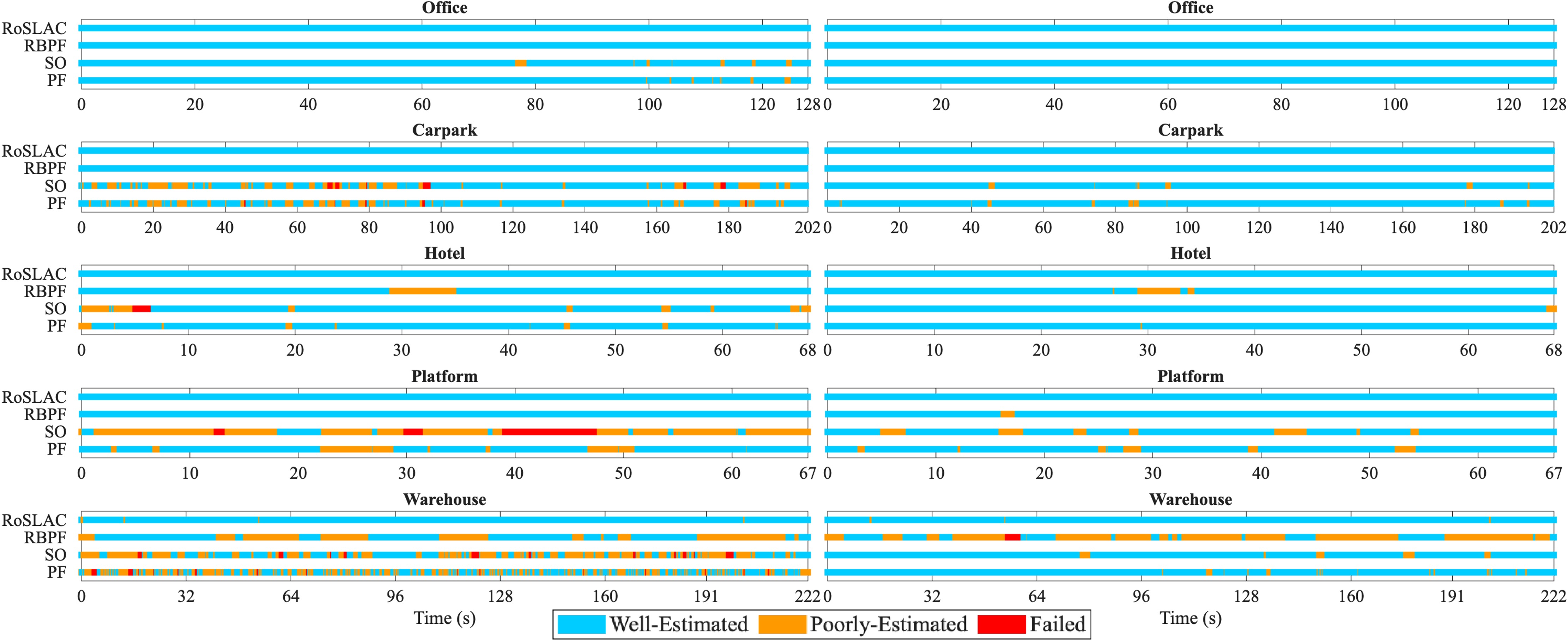}
	\caption{Localization stability in different scenarios. Performance is compared with LiDAR-based relocalization~\cite{shen2025ctemlo}. Errors smaller than $0.3\,\mathrm{m}$ are marked in blue (well estimated), errors between $0.3\,\mathrm{m}$ and $1\,\mathrm{m}$ are marked in orange (poorly estimated), and errors larger than $1\,\mathrm{m}$ are marked in red (failed).}
	\label{fig:exp2}
	\vspace{-1em}
\end{figure*}
\begin{table*}[!t]
	\belowrulesep=0pt
	\aboverulesep=0pt
	\setlength{\tabcolsep}{3pt}
	\caption{Comparison of element-wise calibration parameter accuracy (in $\mu\mathrm{T}$) under different initial guesses, along with the corresponding localization accuracy measured by Absolute Trajectory Error (ATE, in meters).}
	\label{tab:exp5}
	\centering
	\vspace{-1em}
	\renewcommand{\arraystretch}{1.2}
	\resizebox{0.90\textwidth}{!}{%
		\begin{threeparttable}
			\begin{tabular}{lccccccc||cccccccccccccccc}
				\hline\hline
				\multicolumn{1}{c}{} & \multicolumn{1}{c|}{} & \multicolumn{6}{c||}{Calibration Parameters} & \multicolumn{16}{c}{Error} \\ \hline
				\multicolumn{1}{c}{\multirow{2}{*}{Scenario}} & \multicolumn{1}{c|}{\multirow{2}{*}{No.}} & \multirow{2}{*}{$c_{11}$} & \multirow{2}{*}{$c_{22}$} & \multirow{2}{*}{$c_{33}$} & \multirow{2}{*}{$b_{1}$} & \multirow{2}{*}{$b_{2}$} & \multirow{2}{*}{$b_{3}$} & \multicolumn{2}{c}{$c_{11}$} & \multicolumn{2}{c}{$c_{22}$} & \multicolumn{2}{c}{$c_{33}$} & \multicolumn{2}{c}{$b_{1}$} & \multicolumn{2}{c}{$b_{2}$} & \multicolumn{2}{c}{$b_{3}$} & \multicolumn{2}{c}{$\ell_2-norm$} & \multicolumn{2}{c}{ATE} \\ \cmidrule(lr){9-10}\cmidrule(lr){11-12}\cmidrule(lr){13-14}\cmidrule(lr){15-16}\cmidrule(lr){17-18}\cmidrule(lr){19-20}\cmidrule(lr){21-22}\cmidrule(lr){23-24}
				\multicolumn{1}{c}{} & \multicolumn{1}{c|}{} &  &  &  &  &  &  & RS & RP & RS & RP & RS & RP & RS & RP & RS & RP & RS & RP & RS & RP & RS & RP \\ \hline
				\multirow{2}{*}{Office} & \multicolumn{1}{c|}{1} & 1.01 & 0.98 & 0.99 & 19.49 & 20.60 & 20.17 & \textbf{0.02} & 0.03 & \textbf{0.01} & 0.05 & 0.03 & \textbf{0.02} & \textbf{0.64} & 3.47 & \textbf{0.49} & 0.83 & \textbf{0.45} & 3.14 & \textbf{1.06} & 4.82 & \textbf{0.11} & 0.17 \\
				& \multicolumn{1}{c|}{2} & 1.03 & 0.94 & 0.96 & 5.33 & 5.39 & 5.36 & 0.02 & \textbf{0.01} & \textbf{0.01} & 0.02 & 0.03 & \textbf{0.03} & \textbf{0.30} & 0.54 & \textbf{0.17} & 0.46 & \textbf{0.55} & 0.89 & \textbf{0.73} & 1.21 & 0.11 & \textbf{0.10} \\
				\multirow{2}{*}{Carpark} & \multicolumn{1}{c|}{3} & 0.97 & 0.97 & 0.97 & 5.23 & 5.82 & 6.04 & 0.01 & \textbf{0.01} & \textbf{0.01} & 0.04 & \textbf{0.01} & 0.04 & \textbf{0.45} & 0.87 & \textbf{0.40} & 1.03 & \textbf{0.18} & 0.84 & \textbf{0.68} & 1.64 & \textbf{0.10} & 0.20 \\
				& \multicolumn{1}{c|}{4} & 0.98 & 0.98 & 1.00 & 10.85 & 9.87 & 12.68 & \textbf{0.01} & 0.07 & \textbf{0.02} & 0.02 & \textbf{0.01} & 0.03 & \textbf{0.67} & 4.68 & \textbf{0.37} & 1.88 & \textbf{0.20} & 1.19 & \textbf{0.85} & 5.29 & \textbf{0.11} & 0.52 \\
				\multirow{2}{*}{Platform} & \multicolumn{1}{c|}{5} & 0.96 & 0.91 & 0.95 & 5.00 & 4.49 & 5.80 & \textbf{0.03} & 0.08 & \textbf{0.02} & 0.09 & \textbf{0.02} & 0.10 & \textbf{0.51} & 1.87 & \textbf{0.58} & 1.79 & \textbf{0.21} & 2.77 & \textbf{0.93} & 4.50 & \textbf{0.10} & 0.10 \\
				& \multicolumn{1}{c|}{6} & 0.91 & 1.00 & 1.01 & 9.66 & 8.43 & 15.37 & \textbf{0.02} & 0.11 & \textbf{0.01} & 0.10 & \textbf{0.04} & 0.06 & \textbf{0.34} & 3.13 & \textbf{0.45} & 5.02 & \textbf{0.68} & 4.89 & \textbf{0.97} & 8.77 & \textbf{0.10} & 0.17 \\
				\multirow{2}{*}{Warehouse} & \multicolumn{1}{c|}{7} & 0.96 & 0.90 & 0.98 & 8.29 & 7.44 & 7.78 & \textbf{0.01} & 0.01 & 0.01 & \textbf{0.01} & \textbf{0.01} & 0.05 & \textbf{0.90} & 3.82 & \textbf{1.16} & 1.44 & \textbf{0.46} & 2.02 & \textbf{1.56} & 4.57 & \textbf{0.10} & 0.23 \\
				& \multicolumn{1}{c|}{8} & 0.93 & 0.99 & 1.03 & 15.43 & 16.10 & 16.07 & \textbf{0.00} & 0.02 & \textbf{0.00} & 0.01 & \textbf{0.00} & 0.10 & \textbf{0.53} & 7.99 & \textbf{1.62} & 2.74 & \textbf{0.13} & 4.63 & \textbf{1.75} & 9.68 & \textbf{0.12} & 0.30 \\ \hline
				\multicolumn{8}{r||}{Average} & \textbf{0.02} & 0.04 & \textbf{0.01} & 0.04 & \textbf{0.02} & 0.05 & \textbf{0.54} & 3.30 & \textbf{0.65} & 1.90 & \textbf{0.36} & 2.55 & \textbf{1.07} & 5.06 & \textbf{0.10} & 0.22 \\ \hline\hline
			\end{tabular}
			\begin{tablenotes}
				\footnotesize
				\item[1] RS denotes proposed \textit{RoSLAC}, RP denotes RBPF. The best results are highlighted in \textbf{BOLD}.
			\end{tablenotes}
		\end{threeparttable}
	}
	\vspace{-1em}
\end{table*}
We further evaluate the robustness of the proposed method. In addition to wheel odometry, which is used in all compared methods, a global relocalization strategy is applied uniformly for fairness. Specifically, when particle degeneration occurs or the gradient-based solver fails to converge, the real-time ground-truth pose is assigned to the prior state so that localization can resume from a valid estimate at the next step. Fig.~\ref{fig:exp2} shows the robustness of different methods across multiple scenarios. For clearer visualization, three Absolute Trajectory Error (ATE) thresholds are defined: light blue for well-estimated poses ($\mathrm{ATE}<0.3\mathrm{m}$), orange for poorly estimated poses ($0.3\mathrm{m}<\mathrm{ATE}<1.0\mathrm{m}$), and red for failed estimates ($\mathrm{ATE}>1.0\mathrm{m}$). The left column of Fig.~\ref{fig:exp2} uses raw magnetometer measurements without calibration, while the right column uses pre-calibrated measurements.

As shown in Fig.~\ref{fig:exp2}, when pre-calibrated magnetometer data are used, all methods achieve generally good localization performance, with most poses estimated stably and accurately. In contrast, without magnetometer calibration, only RBPF and the proposed method maintain robust localization over the full trajectory, owing to their ability to update calibration parameters online. In several scenarios, the proposed method demonstrates slightly stronger robustness than RBPF which occasionally exhibits short periods of degraded estimation.

\subsubsection{Calibration Robustness}
For robust calibration, the algorithm should converge to the global optimum even when the initial guess is far from it. To further evaluate this property, Table~\ref{tab:exp5} presents results from eight runs with randomly generated calibration parameters. Specifically, magnetometer measurements are first undistorted using pre-calibrated parameters and then re-distorted using random calibration parameters before being used for localization and calibration. In all runs, the identity matrix is adopted as the initial guess. Table~\ref{tab:exp5} reports the diagonal elements of the calibration matrix $\mathbf{C}$ and the bias vector $\mathbf{b}$, averaged over the eight magnetometers for each run. The \textit{RS}/\textit{RP} columns further report the corresponding post-calibration estimation errors, together with the $\ell_2$-norm error of the calibration parameter $\boldsymbol{\theta}$.

The results show that the proposed \textit{RoSLAC} method is largely insensitive to the magnitude of the initial deviation. The estimation of $\mathbf{C}$ remains accurate, with the averaged element-wise error consistently below $0.02$, while the bias estimation is also robust. Even when the initial bias reaches about $27\mu\mathrm{T}$, the final error is reduced to around $1.7\mu\mathrm{T}$, and the average error over all settings is only $1.07\mu\mathrm{T}$. These results demonstrate strong robustness to poor initialization. In contrast, RBPF achieves high calibration accuracy only when initialized close to the ground truth (e.g., \textit{Office-2} and \textit{Carpark-3}), but often fails to converge to the global optimum under large initial deviations (e.g., \textit{Platform-6} and \textit{Warehouse-8}). This limitation likely arises because: (a) the linear calibration parameters in RBPF are tightly coupled with nonlinear state estimation, making the filter prone to premature convergence when the problem is ill-conditioned in the early stage and causing it to lose sensitivity to the global solution; and (b) each measurement is processed only once, which reduces the ability to escape local optima and may introduce bias into both calibration and localization.

The corresponding ATE values are also reported to examine their relationship with calibration accuracy. For both methods, the overall ATE remains relatively small. However, \textit{RoSLAC} maintains stable localization accuracy at about $0.1\mathrm{m}$ across all initializations, consistent with its stable calibration performance. By contrast, larger calibration errors in RBPF are associated with reduced localization stability.
\vspace{-1em}
\subsection{Algorithm Efficiency}
\begin{figure*}[!t]
	\centering
	\includegraphics[width=0.85\linewidth]{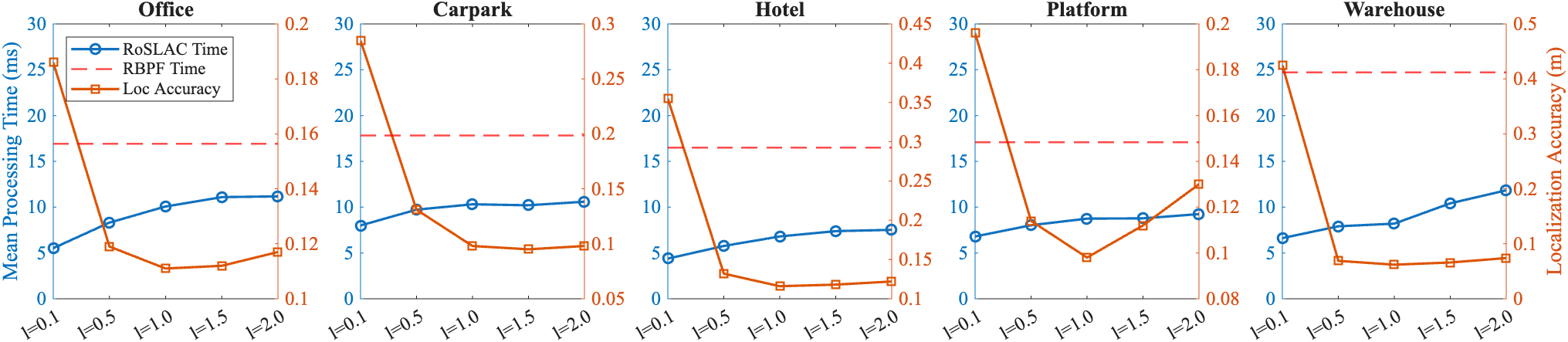}
	\caption{Effect of the sequence accumulation length on the mean per-frame processing time of \textit{RoSLAC} and the corresponding localization accuracy. The mean processing time is additionally compared with that of RBPF as a reference.}
	\label{fig:exp6}
	\vspace{-1em}
\end{figure*}
One key hyperparameter of \textit{RoSLAC} is the accumulated sequence length $l$, which influences localization accuracy, robustness, and real-time performance. To study its effect, we vary $l$ across different scenarios and measure the mean processing time over the full trajectory. The reported time corresponds to calibration of all eight magnetometers at one time step. For comparison, the mean processing time of RBPF is also included, although it does not use sequence accumulation. The localization error of \textit{RoSLAC} for each $l$ is summarized in Fig.~\ref{fig:exp6}.

In Fig.~\ref{fig:exp6}, the horizontal axis denotes the accumulated sequence length in meters. Longer sequences increase the processing time from about $5\mathrm{ms}$ to $10\mathrm{ms}$. For a fixed $l$, however, the runtime is similar across scenarios, indicating that computation time is dominated by optimization rather than map queries. In terms of accuracy, $l=0.5\mathrm{m}$ is sufficient for reliable localization. At this setting, the per-frame processing time is about $7.5\mathrm{ms}$, corresponding to an output rate of roughly $133\mathrm{Hz}$. By contrast, RBPF requires at least $15\mathrm{ms}$ per frame, making it about twice as slow as the proposed method.
\vspace{-1.5em}
\section{Ablation Study}\label{sec:ablation}
To illustrate both the effectiveness and the necessity of each module in the proposed \textit{RoSLAC} framework, we conduct an ablation study by evaluating different combinations of modules and resulting localization accuracy. Specifically, we report the final localization ATE under different scenarios. The evaluated module combinations are summarized in Table~\ref{tab:abl_1}, which includes four distinct settings.

When the magnetometers are not calibrated beforehand, the joint use of all modules becomes critical. As shown in Table~\ref{tab:abl_2}, the proposed full method achieves the smallest ATE of approximately $0.1\mathrm{m}$, whereas removing any individual module leads to a significant degradation in performance, with the largest error approaching $0.7\mathrm{m}$. By comparing S2 and S3, it can be observed that when online calibration is required, incorporating the sequence accumulation module is essential. This is because calibration increases the degrees of freedom of the overall system, and sequence accumulation provides additional observations to sufficiently constrain the estimation problem. Furthermore, comparing S1 and S3, we observe that sequence accumulation can partially compensate for the distortions introduced by calibration. Nevertheless, the performance of S1 remains inferior to that of the full proposed configuration. When the magnetometers are pre-calibrated, the calibration module becomes less critical as expected. In this case, some small ATE values are achieved under S1. However, the pre-calibration is not guaranteed to be perfect, thus the calibration module can provide crucial assistance and greatly improve robustness in practical deployments.

\begin{table}[]
	\centering
	\caption{Different combinations of the modules proposed in \textit{RoSLAC}.}
	\label{tab:abl_1}
	\vspace{-1em}
	\renewcommand{\arraystretch}{1.2}
	\begin{tabular}{r|cccc}
		\hline\hline
		\multicolumn{1}{c|}{Algorithm Block} & \textbf{RoSLAC} & S1 & S2 & S3 \\ \hline
		Sequence Accumulation                & \checkmark                 & \checkmark  &    &    \\
		Online Calibration                   & \checkmark                 &    & \checkmark  &    \\
		Online Localization                  & \checkmark                 & \checkmark  & \checkmark  & \checkmark  \\ \hline\hline
	\end{tabular}
	\vspace{-1em}
\end{table}

\begin{table}[]
	\centering
	\caption{The localization accuracy comparison in Absolute Trajectory Error (ATE, meter). The best result is highlighted in \textbf{BOLD}}
	\label{tab:abl_2}
	\vspace{-1em}
	\renewcommand{\arraystretch}{1.2}
	\begin{adjustbox}{width=0.99\linewidth}
		\begin{tabular}{c|cccccccc}
			\hline\hline
			\multirow{3}{*}{Scenario} & \multicolumn{8}{c}{Localization ATE (m)}                                                                                                                                  \\ \cline{2-9} 
			& \multicolumn{2}{c|}{\textbf{RoSLAC}}               & \multicolumn{2}{c|}{S1}                     & \multicolumn{2}{c|}{S2}                     & \multicolumn{2}{c}{S3} \\ \cline{2-9} 
			& raw            & \multicolumn{1}{c|}{pre-calib}      & raw   & \multicolumn{1}{c|}{pre-calib}      & raw   & \multicolumn{1}{c|}{pre-calib}      & raw      & pre-calib   \\ \hline
			Office                   & \textbf{0.114} & \multicolumn{1}{c|}{\textbf{0.114}} & 0.168 & \multicolumn{1}{c|}{0.119}          & 0.368 & \multicolumn{1}{c|}{0.433}          & 0.232    & 0.130       \\
			Carpark                    & \textbf{0.103} & \multicolumn{1}{c|}{\textbf{0.104}} & 0.330 & \multicolumn{1}{c|}{0.132}          & 0.512 & \multicolumn{1}{c|}{0.547}          & 0.405    & 0.159       \\
			Hotel                      & \textbf{0.117} & \multicolumn{1}{c|}{0.118}          & 0.190 & \multicolumn{1}{c|}{\textbf{0.113}} & 0.504 & \multicolumn{1}{c|}{0.439}          & 0.387    & 0.183       \\
			Platform                   & \textbf{0.093} & \multicolumn{1}{c|}{0.083}          & 0.231 & \multicolumn{1}{c|}{0.088}          & 0.533 & \multicolumn{1}{c|}{\textbf{0.080}} & 0.301    & 0.092       \\
			Warehouse                  & \textbf{0.087} & \multicolumn{1}{c|}{0.065}          & 0.669 & \multicolumn{1}{c|}{\textbf{0.023}} & 0.373 & \multicolumn{1}{c|}{0.294}          & 0.727    & 0.076       \\ \hline\hline
		\end{tabular}
	\end{adjustbox}
	\vspace{-1.5em}
\end{table}

\section{Conclusions and Future Work} \label{sec:conclude}
This paper presented a novel robust simultaneous localization and calibration method, termed \textit{RoSLAC}, for jointly estimating robot pose and magnetometer calibration parameters in a pre-built magnetic map. The framework comprises three key components: a sequence accumulator, an alternating optimization module, and a post-processing filter. To mitigate the limited observability, magnetometer measurements are accumulated along the odometry trajectory to provide more informative constraints. Based on these observations, the alternating optimization module iteratively refines both pose and calibration parameters, while the post-processing filter further drives the solution toward globally optimal calibration. Extensive experiments in both simulation and real-world environments showed that \textit{RoSLAC} can robustly estimate calibration parameters online with an average error of about $1\,\mu\text{T}$, even under poor initialization, while maintaining stable localization with an average ATE of approximately $10\,\text{cm}$. Future work will focus on CUDA-based acceleration of the Gauss-Newton gradient computation to further improve computational efficiency on mobile platforms.

\bibliographystyle{ieeetr}
\bibliographystyle{./bibliography/IEEEtran}
\bibliography{mybib}

 \vspace{-3em}

\begin{IEEEbiography}[{\includegraphics[width=1in,height=1.25in,clip,keepaspectratio]{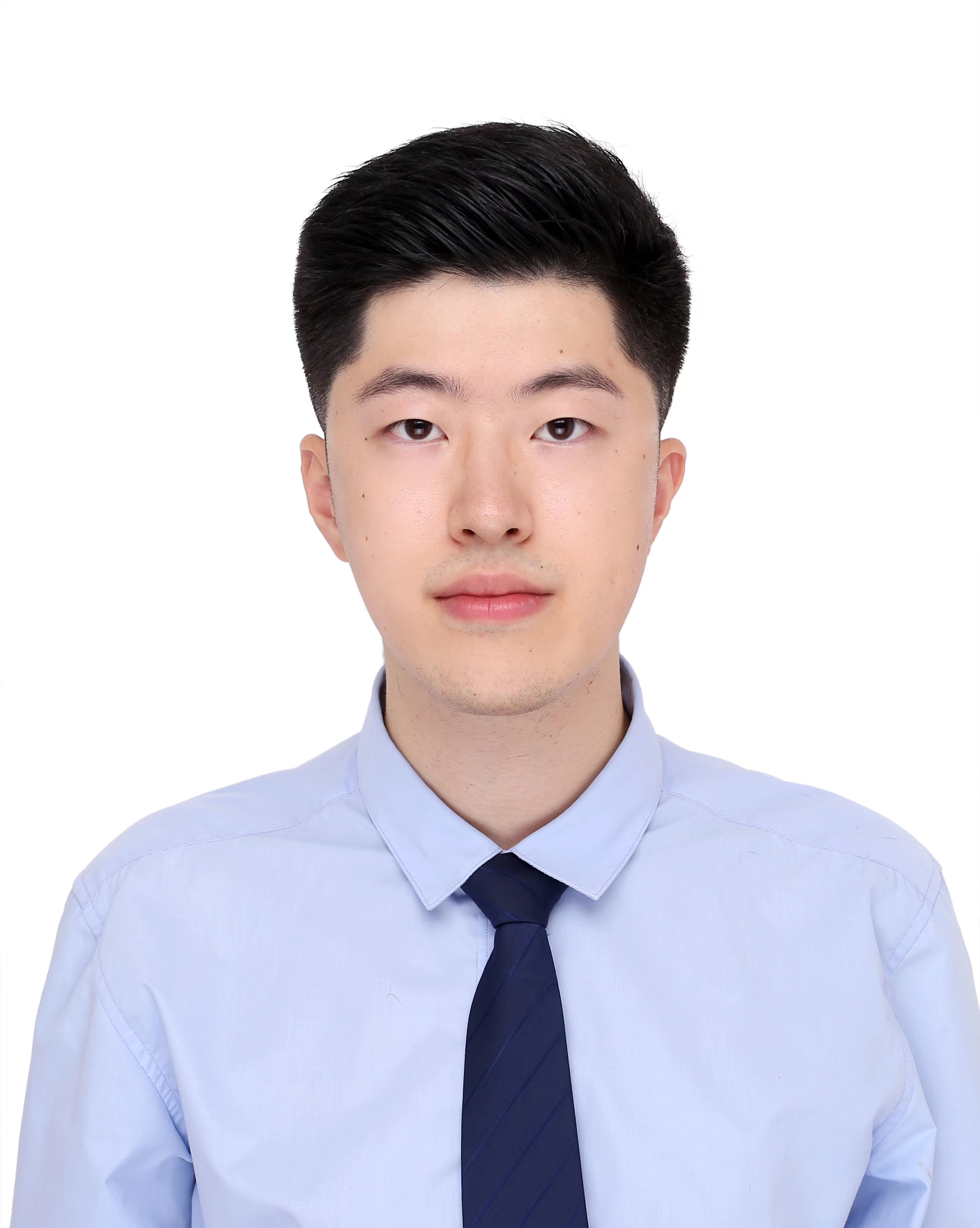}}]{Qiyang Lyu}
	received his B.Eng degree of Electronic Information Engineering from University of Electronic Science and Technology of China, China, in 2020, and the M.Sc. degrees of Computer Control and Automation from Nanyang Technological University, Singapore, in 2021. Now he is pursuing the Ph.D. degree with the School of Electrical and Electronic Engineering, Nanyang Technological University, Singapore. His research interests include sensor calibration, multi-modal mapping, and localization for autonomous robots in real complex scenarios.
\end{IEEEbiography}
\vspace{-3em}
\begin{IEEEbiography}[{\includegraphics[width=1in,height=1.25in,clip,keepaspectratio]{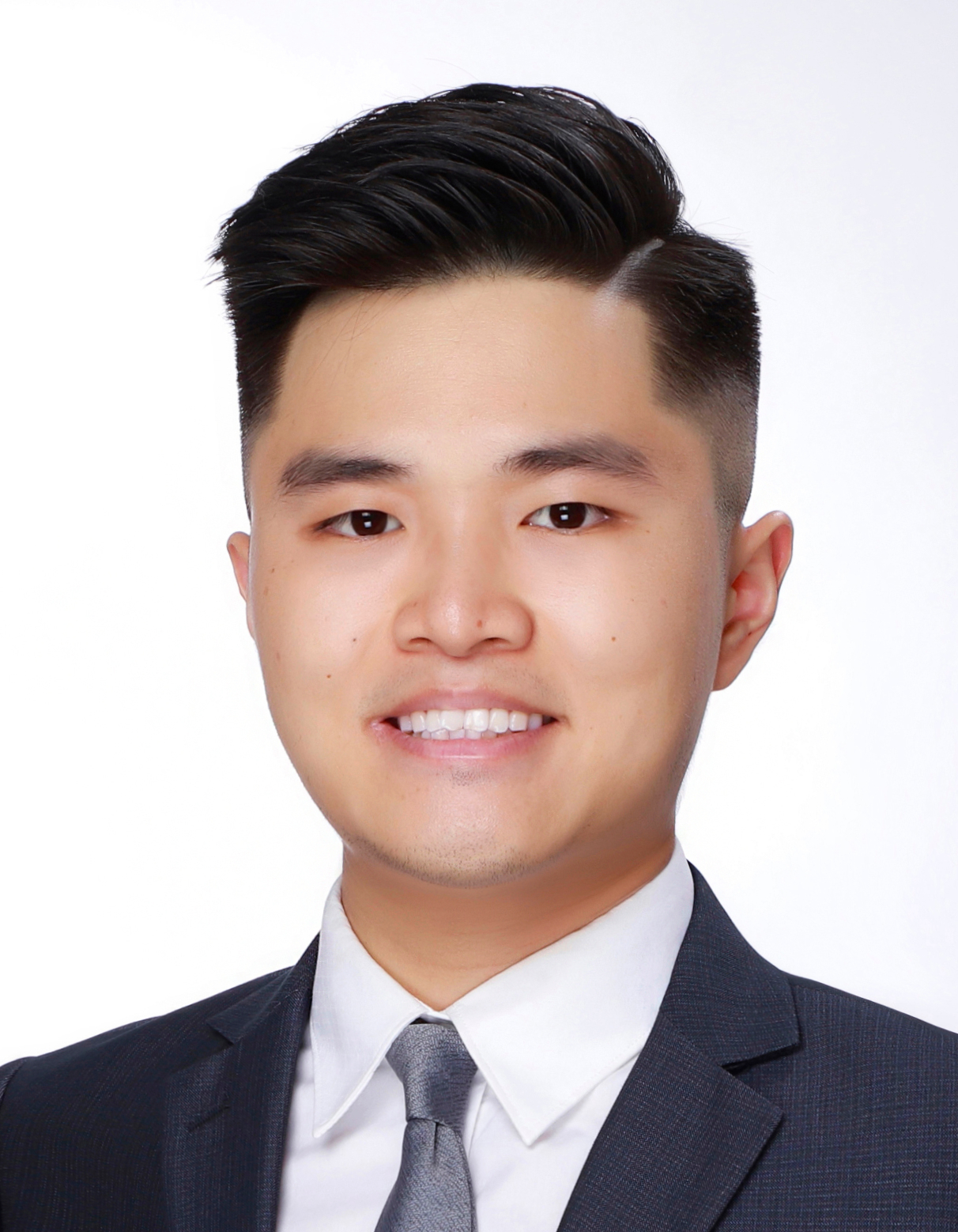}}]{Zhenyu Wu}
	received his B.Eng. degree of Electrical Engineering and Automation from Wuhan University, China, in 2016, the M.Sc. and Ph.D. degrees from Nanyang Technological University (NTU), Singapore, in 2017 and 2022, respectively. He is currently a Research Assistant Professor with the Centre for Advanced Robotics Technology Innovation (CARTIN), NTU. He served as an Associate Editor for the IEEE/RSJ IROS in 2025 and is serving as an editorial board member for Robot Learning. His research interests include intelligent perception, localization, and navigation for autonomous systems in complex environments. 
\end{IEEEbiography}
\vspace{-3em}
\begin{IEEEbiography}[{\includegraphics[width=1in,height=1.25in,clip,keepaspectratio]{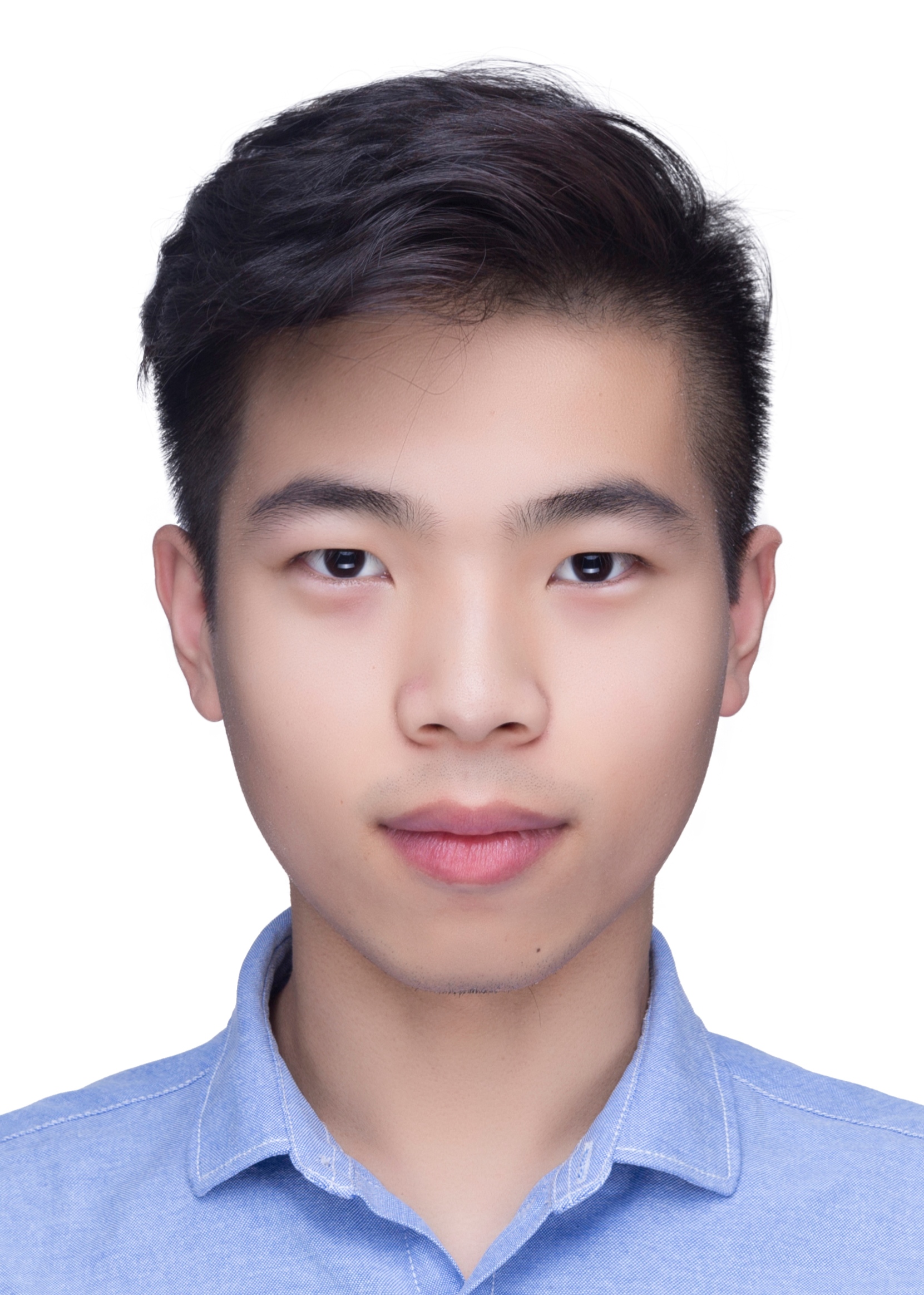}}]{Wei Wang}
	received his B.Eng degree of Automation from Zhejiang University, China, in 2020, and the M.Sc. degrees of Computer Control and Automation from Nanyang Technological University, Singapore, in 2021. Now he is pursuing the Ph.D. degree with the School of Electrical and Electronic Engineering, Nanyang Technological University, Singapore. His research interests include autonomous robots localization and navigation in complex terrains, and intelligent transportation systems.
\end{IEEEbiography}
\vspace{-3em}
\begin{IEEEbiography}[{\includegraphics[width=1in,height=1.25in,clip,keepaspectratio]{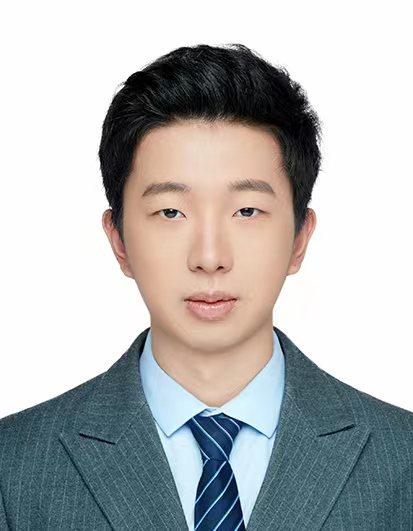}}]{Hongming Shen}
	received the B.S. degree in Flight Vehicle Design and Engineering from Central North University, Taiyuan, China, in 2015, and the M.S. degree in Aerospace Transportation and Control from the Beijing Institute of Technology, Beijing, China, in 2017. He received the Ph.D. degree in Control Theory and Control Engineering from Tianjin University, Tianjin, China, in 2023. He is currently a Postdoctoral Research Fellow with the Centre for Advanced Robotics Technology Innovation (CARTIN), NTU. His current research interests include state estimation, multisensor fusion, localization and mapping, and aerial Robotics.
\end{IEEEbiography}
\vspace{-3em}
\begin{IEEEbiography}[{\includegraphics[width=1in,height=1.25in,clip,keepaspectratio]{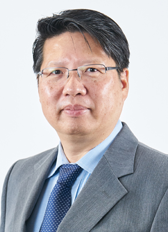}}]{Danwei Wang} (Life Fellow, IEEE) received his Ph.D. and M.S.E. degrees from the University of Michigan, Ann Arbor, USA, in 1989 and 1984, respectively. He received his B.E degree from the South China University of Technology, China, in 1982. Since 1989, he has been with the School of Electrical and Electronic Engineering, Nanyang Technological University, Singapore. Currently, he is Emeritus Professor and was the Director of the ST Engineering-NTU Robotics Corporate Lab. He is the Chair of IEEE Singapore Robotics and Automation Chapter and a senator in NTU Academics Council. He has served as general chairman, technical chairman and various positions in international robotics and control conferences, such as ICRA, IROS, and ICARCV conferences. He was a recipient of Alexander von Humboldt Fellowship, Germany, and ST Engineering Distinguished Professor Award, Singapore. He is a Fellow of the Academy of Engineering, Singapore (SAEng) and Life Fellow of the IEEE. His research interests include robotics, control theory and applications. 
\end{IEEEbiography}

\end{document}